\documentclass[journal]{IEEEtran}
\usepackage{amsmath,amsfonts}
\usepackage{algorithm}
\usepackage{array, multirow, multicol}
\usepackage[caption=false,font=normalsize,labelfont=sf,textfont=sf]{subfig}
\usepackage{textcomp}
\usepackage{stfloats}
\usepackage{url}
\usepackage{verbatim}
\usepackage{graphicx}
\usepackage{cite}
\newtheorem{myDef}{Definition}
\newtheorem{theorem}{Theorem}
\newtheorem{lemma}{Lemma}

\usepackage{algpseudocode}

\algnewcommand\algorithmicinput{\textbf{Input:}}
\algnewcommand\Input{\item[\algorithmicinput]}
\algnewcommand\algorithmicoutput{\textbf{Output:}}
\algnewcommand\Output{\item[\algorithmicoutput]}

\begin{document}

\title{OTAD: An Optimal Transport-Induced Robust Model for Agnostic Adversarial Attack}

\author{Kuo Gai, Sicong Wang, Shihua Zhang
\thanks{The first two authors contributed equally. 
Kuo Gai is with the Shanghai Institute for Mathematics and Interdisciplinary Sciences (SIMIS), Shanghai 200433, China,
Sicong Wang and Shihua Zhang are with the Academy of Mathematics and Systems Science, Chinese Academy of Sciences, Beijing 100190, China, and also with the School of Mathematical Sciences, University of Chinese Academy of Sciences, Beijing 100049, China (Corresponding author: Shihua Zhang, E-mail: zsh@amss.ac.cn).}
}


\maketitle

\begin{abstract}
Deep neural networks (DNNs) are vulnerable to small adversarial perturbations of the inputs, posing a significant challenge to their reliability and robustness. Empirical methods such as adversarial training can defend against particular attacks but remain vulnerable to more powerful attacks. Alternatively, Lipschitz networks provide certified robustness to unseen perturbations but lack sufficient expressive power. To harness the advantages of both approaches, we design a novel two-step Optimal Transport induced Adversarial Defense (OTAD) model that can fit the training data accurately while preserving the local Lipschitz continuity. First, we train a DNN with a regularizer derived from optimal transport theory, yielding a discrete optimal transport map linking data to its features. By leveraging the map's inherent regularity, we interpolate the map by solving the convex integration problem (CIP) to guarantee the local Lipschitz property. OTAD is extensible to diverse architectures of ResNet and Transformer, making it suitable for complex data. For efficient computation, the CIP can be solved through training neural networks. OTAD opens a novel avenue for developing reliable and secure deep learning systems through the regularity of optimal transport maps. Empirical results demonstrate that OTAD can outperform other robust models on diverse datasets. 
\end{abstract}


\section{Introduction}
\IEEEPARstart{D}{eep} neural networks (DNNs) are the most crucial component of the artificial intelligence (AI) field. DNNs are rapidly becoming the state-of-the-art approaches in many tasks, i.e., computer vision, speech recognition, and natural language processing. Theoretical explorations of DNNs inspire the understanding of deep learning and development of new algorithms \cite{tishby2015deep,zhang2021NSR, Zhang_Zhang_2022, wang2024pfc, ruan2024attention, 9779314, 10288226}. However, DNNs are vulnerable to adversarial attacks, i.e., a well-chosen small perturbation of the input data can lead a neural network to predict incorrect classes.

Various strategies have been proposed to enhance the robustness of existing models \cite{10263803}. These strategies include adversarial training \cite{szegedy2013intriguing,goodfellow2014explaining,huang2015learning}, where adversarial examples are generated during training and added to the training set. However, these modified models often exhibit vulnerabilities against strong adversaries \cite{carlini2017adversarial,athalye2018robustness,uesato2018adversarial,athalye2018obfuscated}, primarily because DNNs require large gradients to represent their target functions and attacks can always take advantage of the gradients to construct adversarial examples. To stop playing this cat-and-mouse game, a growing body of studies have focused on certified robustness. 

One straightforward approach to certified robustness involves constraining their Lipschitz constant. Existing approaches to enforce Lipschitz constraints can be categorized into three groups: soft regularization \cite{gulrajani2017improved,drucker1992improving,sokolic2017robust}, hard constraints on weights \cite{yoshida2017spectral,gouk2021regularisation,tsuzuku2018lipschitz,cisse2017parseval} and specifically designed activations \cite{anil2019sorting,zhang2021towards,zhang2021boosting}. However, compared to standard networks, these approaches show suboptimal performance even on simple datasets like CIFAR10. The strict Lipschitz constraints during training may hinder the model's ability to find a more effective Lipschitz function. Additionally, the target function is not Lipschitz everywhere, especially in classification problems on continuous data distributions where the function cannot be Lipschitz at the boundary between two classes.

In this paper, we propose a novel two-step model named OTAD to combine the strengths of the mentioned approaches (Fig. 1). The objective is to achieve a robust and accurate learned function at the terminal stage of training without enforcing Lipschitz constraints throughout the entire training process. Inspired by optimal transport theory, we leverage the theory that the optimal transport map is the derivative of a convex function $\phi$ and possesses regularity properties, implying the map $\nabla \phi$ is locally Lipschitz under moderate conditions. Based on this, we can learn the discrete optimal transport map through neural networks during training and compute the robust output of the model that satisfies the local Lipschitz property. 

In detail, we first employ a DNN to acquire the optimal transport map from data to the feature for classification (Fig. 1). Gai and Zhang \cite{gai2021mathematical} have demonstrated that ResNet with weight decay tends to approximate the Wasserstein geodesics during training. Therefore, we first utilize ResNet to obtain a discrete optimal transport map $T$ from data points to their features. $T$ can accurately classify the training data due to the approximation power of ResNet. Subsequently, we employ a robust model based on the discrete optimal transport map $T$ instead of the learned ResNet. For arbitrary given input $x$ in the inference process, our objective is to find an appropriate feature $y$ such that a Lipschitz function $f$ exists, satisfying $f$ being consistent with the discrete optimal transport map on the training set and $f(x)=y$. Given a set $\{(x_i, T(x_i))\}_{i\in I}$, the goal is to find a convex and smooth function $g$ such that $\nabla g(x_i)=T(x_i)$. This problem can be formalized into a convex integration problem (CIP). We demonstrate that solving a quadratically constrained program (QCP) based on recent advances in first-order methods \cite{taylor2017convex} can find a solution to the CIP and yield a feasible value of $y$. 

However, the QCP is much slower than the inference of a DNN. To address this issue, we train a Transformer named CIP-net as an alternative to the optimization algorithm for efficient computation. Theoretically, we derive an upper bound of the Lipschitz constant of the Transformer block, demonstrating the strong performance of CIP-net.

To further improve the performance of OTAD, we extend it with various architectures and metric learning. 
As Transformers employ residue connections in forward propagation, and the invariant dimension of the feature aligns well with the optimal transport setting, we adapt OTAD to Transformer-based architecture such as ViT \cite{dosovitskiy2021an}. 
Finding neighbors is a critical step in OTAD, and the $l_2$ distance may not effectively characterize the similarity of data closing to a manifold in high dimensional space. Consequently, a more general version of OTAD finds neighbors with learnable metrics. We explore metric learning to find more suitable neighbors and explore the trade-off between accuracy and vulnerability in such cases. 
Besides, we can randomly choose a training subset for the neighborhood search process to reduce the memory and computational cost. Thus, OTAD can be scaled to large complex datasets like ImageNet.
We implement various experiments to test the proposed OTAD model and its variants under different settings. Empirical results demonstrate superior performance to adversarial training methods and Lipschitz networks across diverse datasets.

The rest of this paper is organized as follows. In section 2, we introduce some related works about adversarial defense. In section 3, we present the background of our method: the optimal transport theory and regularity of the optimal transport map. In section 4, we develop the Optimal Transport-based Adversarial Defense model (OTAD) and its implementation details. In section 5, we perform extensive experiments to demonstrate the defense ability of OTAD. 

\begin{figure*}[!t]
\centering
\includegraphics[width=0.90\textwidth]{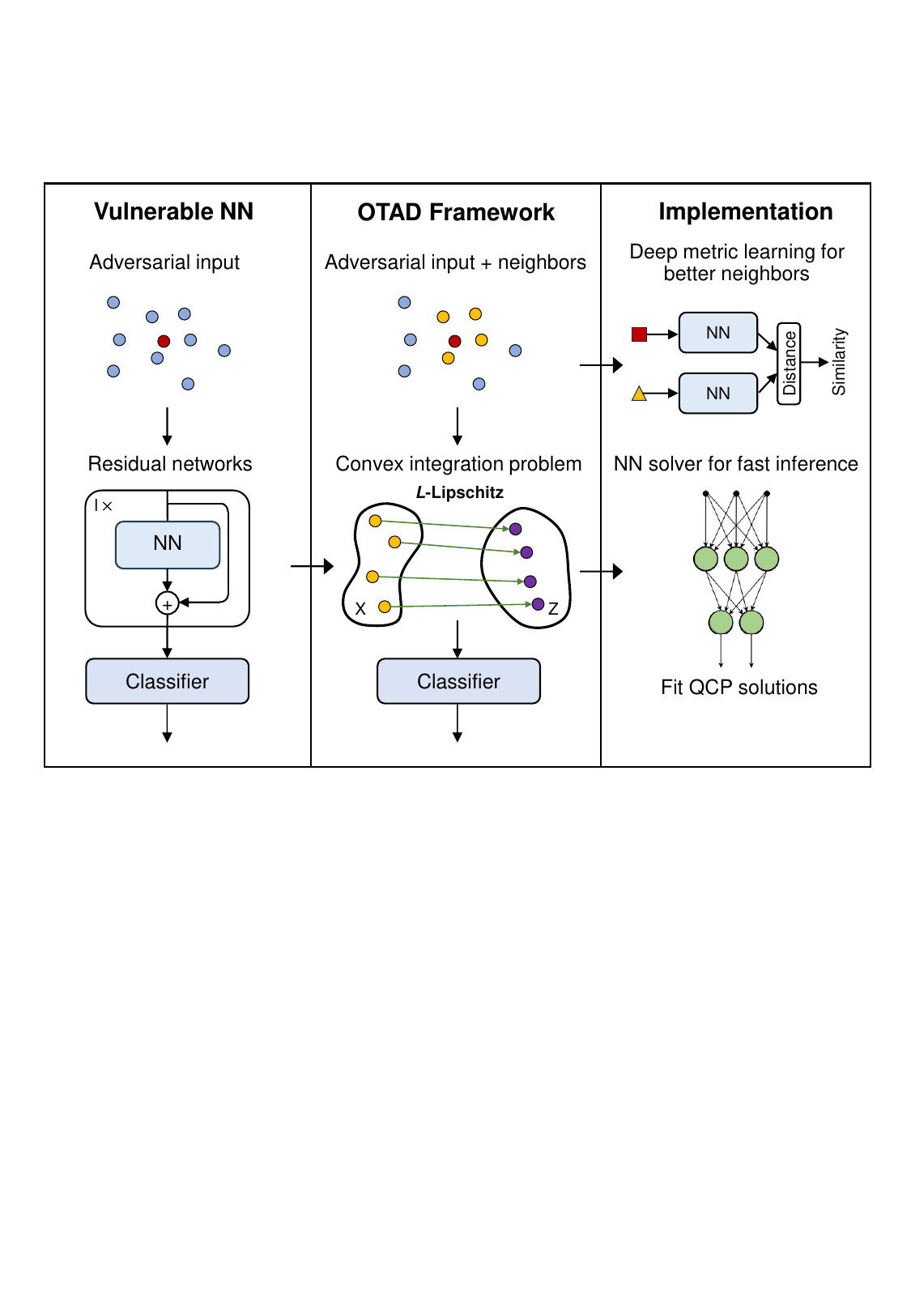}
\caption{\textbf{Illustration of OTAD and its implementation.} DNNs are vulnerable to small adversarial perturbations of the inputs. To classify the adversarial inputs accurately, OTAD replaces the inference process of DNN by solving a convex integration problem with a neighborhood set, which guarantees the local Lipschitz property. Furthermore, OTAD can adopt deep metric learning to find more similar neighborhood sets. For fast inference, the CIP can be solved by employing a neural network trained with the solution of the QCP problem. }
\label{figure1}
\end{figure*}

\section{Related works}
\subsection{Adversarial training}
\par Adversarial training aims at resisting adversarial attacks by worst-case risk minimization. Consider a classifier neural network $g$ with trainable parameters $\theta$, let $\mathcal{L}(g(x),y)$ denote the  classification loss on data $x$ and its label $y$, then the objective of $g$ is
\begin{equation}
	\min_\theta \mathbb{E}_{p(x,y)}\left[\max_{x^\prime \in \mathcal{B}(x)} \mathcal{L}(g(x^\prime),y)\right]
\end{equation}
where $p(x,y)$ is the joint distribution of data and labels, $\mathcal{B}(x)$ is a neighborhood of $x$. However, it is impossible to search the whole region since the loss surface of networks is complex. To approximate it, adversarial training methods \cite{szegedy2013intriguing,goodfellow2014explaining,huang2015learning,kurakin2016adversarial,madry2017towards,ding2018mma,zhang2019theoretically,wong2020fast,Wang2020Improving,wu2020adversarial} add adversarial examples (often found by gradient descent) to the training set during training. After training, the model is robust to the type of attack chosen in the training process. The defense can still be `broken' by stronger adversaries \cite{carlini2017adversarial,athalye2018robustness,uesato2018adversarial,athalye2018obfuscated}. To escape the cat-and-mouse game, training neural networks with bounded Lipschitz constant has been considered a promising way to defend against attacks.  

\subsection{Adversarial purification}
\par Adversarial purification aims at reconstructing the clean data point by conventional transform \cite{xu2017feature,sun2019adversarial} or generative model before classification  \cite{shi2021online,yoon2021adversarial}. Let $P$ be the distribution of clean data. For adversarial noise corrupted data point $x+n_{ad}$, generative model-based adversarial purification trains a network $G$ to project the noisy data back to the manifold of clean data distribution, i.e., $G(x+n_{ad})=x$. The classifier $C$ is trained on $P$ and will classify $G(x+n_{ad})$ with better accuracy. 

Adversarial purification can defend unseen attacks because the generative model $G$ is trained independently from adversarial attacks $n_{ad}$ and classifiers $C$. However, we can still construct adversarial examples by taking the gradients of the generative process. Such adversarial noise will not be purified \cite{yoon2021adversarial}. The effectiveness of adversarial purification methods is highly related to the performance of generative models, such as energy-based models (EBM)\cite{du2019implicit}, generative adversarial networks (GAN) \cite{samangouei2018defensegan,jin2019ape} and auto-regressive generative models\cite{song2017pixeldefend}. The diffusion model is a rising effective generative model. Diffusion-based adversarial methods have achieved competitive performance on large-scale image datasets \cite{DBLP:conf/icml/NieGHXVA22,silva2023diffdefense}. For small datasets (e.g., single-cell gene expression data) with specific noise or missing values, it can be hard to train an effective generative model, indicating that adversarial purification may not be a general method for various datasets.
 
\subsection{Lipschitz networks and random smoothing}
Training neural networks under a Lipschitz constraint puts a bound on how much its output can change in proportion to a change in its input. Existing methods fall into three categories: soft regularization, hard constraints for weights, and specifically designed activations. 

\par Regularizations such as penalizing the Jacobian of the network \cite{gulrajani2017improved,drucker1992improving,sokolic2017robust} can constrain the Lipschitz condition along some directions but does not provably enforce a local Lipschitz constraint on a $\epsilon$-ball of a data point. Thus, adding such regularizations cannot solve the fragility of adversarial attacks on neural networks.

\par On the other hand, as 1-Lipschitz functions are closed under composition, it suffices to constrain 1-Lipschitz affine transformations and activations. Several prior works \cite{yoshida2017spectral,gouk2021regularisation,tsuzuku2018lipschitz} enforce the Lipschitz property by constraining the spectral norm of each weight matrix to be less than one. Another method \cite{cisse2017parseval} projects the weights closer to the manifold of orthogonal matrices after each update. These models provably satisfy the Lipschitz constraint but lack expressivity to some simple Lipschitz functions. 

\par To enhance the expressivity of the Lipschitz network, Anil \textit{et al.} \cite{anil2019sorting} proposes a gradient norm-preserving activation function named GroupSort and proves that the networks with this activation and matrix-norm constraints are universal approximators of Lipschitz function. Singla \textit{et al.} \cite{singla2021improved} further provides some tricks such as certificate regularization to boost the robustness. Zhang \textit{et al.} \cite{zhang2021towards,zhang2021boosting} propose to use $l_\infty$-distance function which is also proved to have universal Lipschitz function approximation property. Though many improvements have been made in this direction, the performance of the above Lipschitz networks is still not satisfactory even on simple datasets like CIFAR10, partially because the strict Lipschitz constraint in the training process impedes the model from finding a better Lipschitz function. Fundamentally different from pursuing Lipschitz constraints in the training process, we propose a novel model that provides robustness by regularity property of optimal transport map while employing powerful architectures like ResNets and Transformers.

Randomized smoothing is another way to obtain a (probabilistic) certified robustness guarantee. This technique uses a Gaussian smoothed classifier, which predicts the most likely label when Gaussian noise is added to the input of the original classifier. Conversely, OTAD integrates neighborhood information but finds robust output satisfying local Lipschitz properties instead of using the original classifier, which may be vulnerable to attacks.

\section{Method background}
\subsection{Optimal transport}
Optimal transport theory provides valuable tools for quantifying the closeness between probability measures, even when their supports do not overlap. $\mathcal{P}_2(\mathbb{R}^d)$ denotes the set of Borel probability measures with finite second-order moment. For two probability measures $\mu$,$\nu\in \mathcal{P}_2(\mathbb{R}^d)$, let $\Pi(\mu, \nu)$ denote the set of all joint distributions $\pi(x,y)$ whose marginals are $\mu$ and $\nu$ respectively. The Wasserstein distance is defined as the solution of the Kantorovich problem \cite{villani2009optimal}:
\begin{equation}
W_2(\mu,\nu):=\left(\inf_{\pi\in\Pi(\mu,\nu)}\int \|x-y\|_2^2d\pi(x,y)\right)^{1/2}
\end{equation}

In the Monge formulation, maps are considered instead of joint distributions. The Borel map $T$ pushes forward $\mu$ to $\nu$, i.e., $T_\#\mu=\nu$. For any set $A\subset \mathbb{R}^d$, $T_\#\mu(A)=\mu(T^{-1}(A))$. If the feasible map $T$ exists, the Monge formulation is equivalent to the Kantorovich formulation:
\begin{equation}
W_2(\mu,\nu)=\left(\inf_{T:T_\#\mu=\nu}\int \|x-T(x)\|_2^2d\mu(x)\right)^{1/2}
\end{equation}

The Brenier theorem \cite{brenier1991polar} asserts that if $\mu$ is absolutely continuous, there always exists a convex function $\phi$ such that $\nabla \phi_{\#}\mu =\nu$ and $\nabla \phi$ is the optimal transport map sending $\mu$ to $\nu$. This convex function $\phi$ is called a Brenier potential between $\mu$ and $\nu$. 

Consider a constant speed geodesic $(\mu_t)$ on $\mathcal{P}_2(\mathbb{R}^d)$ induced by an optimal transport map $T$ connecting $\mu$ and $\nu$:
\begin{equation}
\mu_t=\left(\left(1-t\right)I+tT\right)_\# \mu, \quad \mu_1=\nu
\end{equation}
The continuity equation for $\mu_t$ is given by:
\begin{equation}
\frac{d}{dt}\mu_t+\nabla \cdot (v_t\mu_t)=0
\end{equation}
where $v_t$ is a vector field on $\mathbb{R}^d$. By Benamou-Brenier formula in \cite{benamou2004numerical}, the constant speed geodesic can be recovered by minimizing the following energy function leading to the third definition of the Wasserstein distance
\begin{equation}\label{eq:4}
W_2(\mu,\nu)=\left(\inf_{\mu_t,v_t} \int_{0}^{1}\|v_t\|_{L^2(\mu_t)}^2dt\right)^{1/2}
\end{equation}
Here, the infimum is taken among all solutions $(\mu_t,v_t)$ satisfying continuity with $\mu_0=\mu$ and $\mu_1=\nu$.

\subsection{Regularity of OT maps}
Regularity in optimal transport is usually understood as the property that the map $\nabla \phi$ is $L$-Lipschitz, equivalent to $\phi$ being $L$-smooth. Assume $\mu$ and $\nu$ are supported on a bounded open set with their density function on the support set bounded away from zero and infinity. When the target domain is convex, Caffarelli \cite{caffarelli1991some} proved that the Brenier map can be guaranteed locally Lipschitz. The optimal transport map exhibits discontinuities at singularities when the target domain is non-convex. Nevertheless, the map remains locally Lipschitz on the support set excluding a Lebesgue negligible set \cite{de2015partial}.

In this paper, we consider regularity (smoothness) and curvature (strong convexity) as the conditions that must be enforced when computing the optimal transport map. Our objective is to find a potential function $\phi$ that is $l$-strongly convex and $L$-smooth, i.e.,
\begin{equation}
l\|x-y\|\leq \|\nabla \phi(x)-\nabla \phi(y)\|\leq L\|x-y\|
\end{equation}

\section{OTAD}
Let $f$ denote the learned function at the terminal stage of the training process, $f^*$ denote the target function that maps the training data points to their labels and $\{x_1,\cdots,x_n\}$ denote the training data points i.i.d. sampled from distribution $P_x$. The primary objective of robust learning is to ensure that the learned function $f$ simultaneously achieves the following two goals:
\begin{itemize}
\item Well-approximation of the target function on the training data, i.e., the loss function 
\begin{equation}
\sum_{i=1}^{n} \mathcal{L}\left(f(x_i),f^*(x_i)\right)
\end{equation}
is minimized.
\item Control over the change in output under input perturbations. That requires the learned function is Lipschitz continuous with a small constant $L$. For any pairs of inputs $x_1$, $x_2 \in \mathbb{R}^d$, the Lipschitz continuity condition is given by:
\begin{equation}
\|f(x_1)-f(x_2)\|\leq L\|x_1-x_2\|
\end{equation}
\end{itemize}
OTAD utilizes DNNs with residual connections to approximate the target function on the training data and enforces the local Lipschitz property by solving a CIP problem. 

\subsection{ResNet-based OTAD}
Consider the classification problem, we first train a $m$-block ResNet $R(\cdot)$ and a classifier $H(\cdot)$ to classify the dataset $\{x_i\}_{i=1}^n$ with labels $\{y_i\}_{i=1}^n$. Given input $x$, the forward propagation of the $k$-th block of $R$ is defined as
\begin{equation}
	x^k=x^{k-1}+R_k(x^{k-1}), \quad k\in \{1,2,\cdots,m\}, x^0=x
\end{equation}
where $R_k$ denotes the shallow network inside the $k$-th block of $R$. Let $R(x)$ be the last-layer features, i.e., $R(x)=x^m$.

The forward Euler discretization of an ODE 
\begin{equation*}
\frac{d x}{d t} = v(x,t)
\end{equation*}
is
\begin{equation*}
x^{k+1}=x^{k}+\Delta t \cdot v(x^{k},t),
\end{equation*}
which is similar to the forward propagation formula of ResNet, omitting the time step $\Delta t$. Thus, ResNet can be viewed as a discretization of an ODE. Furthermore, given data points $\{x_1,x_2,\cdots,x_n\}$, each point has a measure of $\frac{1}{n}$, the ResNet outputs the same number of points as input. This property is measure preserving. An ODE holding measure preserving property satisfies the continuity equation
\begin{equation}
	\frac{\partial p_t}{\partial t}+ \nabla \cdot (v_t p_t)=0
\end{equation}
where $p_t$ is the distribution of $x$ at time $t$ and $v_t (x)=v(x,t)$.

If the initial distribution $p_0$ and the terminal distribution $p_1$ are fixed, infinite ODEs satisfying the boundary condition and continuity equation. The Benamou-Brenier formula said, that the Wasserstein geodesic curve induced by optimal transport map can be recovered by minimizing the energy
\begin{equation}\label{eq:regu}
	\int_{0}^{1}\|v_t\|_{L^2(p_t)}dt
\end{equation}
As discussed above, ResNet can be viewed as a discretization of the ODE satisfying continuity equation. The energy above also has a discrete version to minimize for ResNet, which is
\begin{equation}
	\sum_{i=1}^{n}\sum_{k=1}^{m} \|R_k(x_i^{k-1})\|_2^2
\end{equation}

Using the energy as a regularizer, the objective to train the ResNet $R(\cdot)$ and the linear classifier $H(\cdot)$ is 
\begin{equation}\label{eq:1}
\begin{aligned}
&\min_\theta \sum_{i=1}^{n}\mathcal{L}(H(R(x_i)),y_i)+ \alpha\sum_{i=1}^{n}\sum_{k=1}^{m} \|R_k(x_i^{k-1})\|_2^2\\
&\mbox{s.t.} \quad x_i^k=x_i^{k-1}+R_k(x_i^{k-1}), \quad k\in \{1,2,\cdots,m\}, x_i^0=x_i
\end{aligned}
\end{equation}
where $\mathcal{L}$ is the loss function for classification, $\theta$ denotes all the trainable parameters in $H(R(\cdot))$, and $\alpha$ is the hyperparameter to balance the two terms. Gai and Zhang \cite{gai2021mathematical} demonstrate that the weight decay operation plays a similar role with the regularizer in (\ref{eq:1}) for a Wasserstein geodesic. Let $z_i=R(x_i)$. If the problem $(\ref{eq:1})$ is well solved, then we obtain a network-based classifier $H(R(\cdot))$ with high training accuracy and a discrete optimal transport map from the data $x_i$ to feature $z_i$. 

\par Though the network $R(\cdot)$ learns the discrete optimal transport map $T$, it can still be vulnerable to small perturbations. Therefore, we need to find a more robust function $\tilde{f}$ with $\tilde{f}(x_i)=z_i$. Let $h$ denote the potential function of $\tilde{f}$, i.e., $\tilde{f}(\cdot)=\nabla h(\cdot)$. Since the optimal transport map is locally Lipschitz and has singularities when the target domain is non-convex, then for a given test data point $x^\prime$, we can only trust training data near $x^\prime$ to constrain $\tilde{f}(x^\prime)$. Let $N_K(x^\prime)$ be the set of $K$ nearest neighbors of $x^\prime$ from the training dataset. Assume $h$ is $l$-strongly convex $L$-smooth on the neighborhood of $x^\prime$, then finding appropriate value $\nabla h(x^\prime)$ can be formulated to a convex integration problem:
\begin{myDef}
Let $I$ be a finite index set and $\mathcal{F}_{l,L}$ denote the class of $l$-strongly convex and $L$-smooth function on $\mathbb{R}^d$. Given a set $S=\{(x_i,z_i)\}_{i\in I}$, the $\mathcal{F}_{l,L}$ convex integration problem is finding a function $f\in \mathcal{F}_{l,L}$, such that $z_i=\nabla f(x_i)$ for all $i\in I$. 
\end{myDef}

First, we need to determine the existence of such a function with respect to $N_K(x^\prime)$. We introduce the definition of $\mathcal{F}_{l,L}$-integrable in \cite{taylor2017convex}.
\begin{myDef}
Consider the set $S=\{(x_i,z_i)\}_{i\in I}$. The set $S$ is $\mathcal{F}_{l,L}$-integrable if and only if there exists a function $f\in \mathcal{F}_{l,L}$ such that $z_i=\nabla f(x_i)$ for all $i\in I$. 
\end{myDef}
According to Theorem 3.8 in Taylor \cite{taylor2017convex}, if the set $\{(x,z)| x\in N_K(x^\prime), z=T(x)\}$ is $\mathcal{F}_{l,L}$-integrable, testing equals to find feasible values of $u_i:=h(x_i)$ satisfying
\begin{equation}\label{eq:2}
\begin{aligned}
&\forall x_i,x_j\in N_K(x^\prime),\quad u_i \geq u_{j}+\left\langle z_{j}, x_i-x_{j}\right\rangle
+\frac{1}{2(1-\ell / L)}\cdot \\
&\left(\frac{1}{L}\left\|z_i-z_{j}\right\|^{2}
+\ell\left\|x_i-x_{j}\right\|^{2}-2 \frac{\ell}{L}\left\langle z_{j}-z_i, x_{j}-x_i\right\rangle\right)
\end{aligned}
\end{equation} 
If $\{(x,z)| x\in N_K(x^\prime), z=T(x)\}$ is $\mathcal{F}_{l,L}$-integrable, let $h$ be the desire $l$-strongly convex $L$-smooth function. For test data $x^\prime$, we can find a feasible value of $v=h(x^\prime)$ and $z^\prime=\nabla h(x^\prime)$ by solving the following QCP problem (Theorem 3.14 in \cite{taylor2017convex})
\begin{equation}\label{eq:3}
\begin{aligned}
&\min _{v \in \mathbb{R}, z^\prime \in \mathbb{R}^{d}} v\\
\text { s.t. } &\forall x_i\in N_K(x^\prime), v \geq u_{i}+\left\langle z_{i}, x^\prime-x_{i}\right\rangle\\
&+\frac{1}{2(1-\ell / L)}\left(\frac{1}{L}\left\|z^\prime-z_{i}\right\|^{2}
+\ell\left\|x^\prime-x_{i}\right\|^{2}\right.\\
&\left. -2 \frac{\ell}{L}\left\langle z_{i}-z^\prime, x_{i}-x^\prime\right\rangle\right)
\end{aligned}
\end{equation}

Then we obtain the feature $z^\prime$ used to classify $x^\prime$. If $\{(x,z)| x\in N_K(x^\prime), z=T(x)\}$ is not $\mathcal{F}_{l,L}$-integrable, we repeat the procedures above with smaller $l$ and larger $L$ until we find feasible values. 
Since the smoothness of Brenier potential cannot be guaranteed, the constant $L$ could be very large. To avoid $L$ larger than tolerance, we detect which constraint is violated in inequalities (\ref{eq:2}) for current $L$ and delete or substitute the $x_j$ corresponding to that inequality. By doing this, we can control the $L$ within tolerance.
The training and testing scheme of ResNet-based OTAD is summarized in Algorithms 1 and 2. We use the optimization tool MOSEK \cite{mosek} to find $\{u_i\}_{i=1}^K$ satisfying (\ref{eq:2}) and $z^\prime$ by solving (\ref{eq:3}). 

The Simplex method for linear programming in (\ref{eq:2}) is NP-hard, but since the simplex method works well in most cases and the number of variables in (\ref{eq:2}) is $K$ (usually set $K=5$ or $K=10$), it doesn’t consume too much time. Solving the convex QCP in (\ref{eq:3}) consumes most of the time. The complexity of QCP in (\ref{eq:3}) is $O((K+d)^3)$, where $K$ is the number of neighbors, and $d$ is the dimension of latent space.

\begin{table}[t!]
\begin{minipage}{\columnwidth}
\begin{algorithm}[H]
\begin{algorithmic}[1]\caption{\textbf{ResNet-based OTAD Training}}\label{algo:ricim_train}
\Input data $\{x_i\}_{i=1}^n$, labels $\{y_i\}_{i=1}^n$, hyperparameter $\alpha$
\Output features $\{z_i\}_{i=1}^n$, classifier $H(\cdot)$, ResNet $R(\cdot)$
\Repeat
\State minimize $$\sum_{i=1}^{n}\mathcal{L}(H(R(x_i)),y_i)+ \alpha \sum_{i=1}^{n}\sum_{k=1}^{m} \|R_k(x_i^{k-1})\|_2^2$$ \quad \quad by gradient descent
\Until{convergence}
\State Let $z_i=R(x_i)$, $i\in\{1,2,\cdots,n\}$
\end{algorithmic}
\end{algorithm}
\end{minipage}
\end{table}

\begin{table}[t!]
\begin{minipage}{\columnwidth}
\begin{algorithm}[H]
\begin{algorithmic}[1]\caption{\textbf{ResNet-based OTAD Testing}}\label{algo:ricim_test}
\Input data $\{x_i\}_{i=1}^n$, features $\{z_i\}_{i=1}^n$, test data $x^\prime$, constant $l$, $L$ and $K$, stepsize $\delta^1$, $\delta^2$, classifier $H(\cdot)$ 
\Output predict feature $z^\prime$ and label $y^\prime$
\State $L_t = L$, $l_t = l$
\While{1}
\State Compute the set $N_K(x^\prime)$ by the $l_2$ distance between $x^\prime$ and $x_i$
\State Find feasible Brenier potential values $\{u_i\}_{i=1}^K$ on $N_K(x^\prime)$ satisfying inequalities (\ref{eq:2}) with $L_t, l_t$
\If{$\{u_i\}_{i=1}^K$ exists}
\State Compute $z^\prime$ by solving the problem (\ref{eq:3}) with $L, l$
\State \textbf{break}
\Else
\State $L_t=L_t+\delta^1$, $l_t=l_t-\delta^2$
\EndIf
\EndWhile
\State Let $y^\prime=H(z^\prime)$
\end{algorithmic}
\end{algorithm}
\end{minipage}
\end{table}

\subsection{Transformer-based OTAD}
ResNet-based OTAD requires training an $m$-block dimension-invariant ResNet. Although dimension-invariant ResNets can be easily implemented for various data and tasks, their fixed dimensionality throughout the forward propagation limits their expressive power. To address this, we extend OTAD to the popular Transformer architecture \cite{vaswani2017attention}, named OTAD-T. Unlike ResNets, Transformers embed the input into a high-dimensional space, and subsequent Transformer blocks maintain this dimensionality, resulting in a model with good expressive power while keeping the dimensionality unchanged. 

Due to the residual connections in the Transformer, the forward propagation of the Transformer can also be viewed as a discretization of a continuity equation. Thus, it approximates the Wasserstein geodesic curve induced by the optimal transport map under the regularizer of (\ref{eq:regu}) at the terminal phase of training. One can use the discrete optimal transport map learned by the Transformer and do the same test procedure as ResNet-based OTAD.

In this paper, we focus on the Vision Transformer (ViT) \cite{dosovitskiy2021an}. Given an input image $x_i$, ViT first divides the input image into $N$ non-overlapping patches $\{x_{i,p}\}_{p\in N}$, then each patch undergoes linear embedding $E(\cdot)$, where it is projected into a higher-dimensional space. The sequence of patch embeddings $\{E(x_{i,p})\}_{p\in N}$ is added positional information by an position operator $P(\cdot)$, resulting $\{P(E(x_{i,p}))\}_{p\in N}$. The embeddings of patches $\{P(E(x_{i,p}))\}_{p\in N}$ are fed into the standard Transformer blocks and obtain features $\{z_{i,p}\}_{p\in N}$. Let 
\begin{equation}
    \tilde{x}_i = 
\begin{pmatrix}
P(E(x_{i,1})) \\
P(E(x_{i,2})) \\
\vdots \\
P(E(x_{i,N}))
\end{pmatrix},
\quad 
\tilde{z}_i = 
\begin{pmatrix}
z_{i,1} \\
z_{i,2} \\
\vdots \\
z_{i,N}
\end{pmatrix}.
\end{equation}
The forward propagation of the $k$-th block of ViT is 
\begin{equation}
\begin{aligned}
    \tilde{x}_i^{k_{\text{Attn}}} &= \tilde{x}_i^{k-1}+\operatorname{Attn}_k(\tilde{x}_i^{k-1})\\
    \tilde{x}_i^{k} &= \tilde{x}_i^{k_{\text{Attn}}}+\operatorname{MLP}_k(\tilde{x}_i^{k_{\text{Attn}}})
\end{aligned}
\end{equation}
where $\operatorname{Attn}_k$ and $\operatorname{MLP}_k$ denote the attention and MLP block in the $k$-th block of ViT. The forward propagation of the Transformer can also be viewed as a discretization of a geodesic curve in Wasserstein space.
The corresponding discrete optimal transport map transforms $\{\tilde{x}_i\}_{i=1}^n$ into the feature $\{\tilde{z}_i\}_{i=1}^n$. Then for given test data $x^\prime$, we can embed it and search a neighborhood set in $\{\tilde{x}_i\}_{i=1}^n$. Finally, we estimate an output feature through solving a problem analogous to (\ref{eq:3}).

\subsection{Neural network for CIP}
In the procedure of OTAD, the most time-consuming step is solving the CIP, particularly the QCP problem for the feature of test data. Traditional QCP solvers like MOSEK \cite{mosek} can be time-consuming when solving large-scale QCPs.

Recently, neural networks have been applied to various optimization problems \cite{bengio2021machine,chen2022learning}, achieving faster solving speeds and even higher accuracy than traditional solvers. Neural networks can efficiently capture complex patterns and dependencies in data, making them well-suited for high-dimensional and large-scale optimization tasks. To improve the inference speed of OTAD-T, we designed an end-to-end neural network to replace traditional solvers for solving the entire convex integration problem. This method is called OTAD-T-NN. 

Specifically, the inputs of CIP are the embeddings of test data $\tilde{x}$, its neighbors in training set $N_K(\tilde{x})$ and the features corresponding to the neighbors $Z(\tilde{x})=\{\tilde{z}_i|\tilde{x}_i \in N_K(\tilde{x})\}$. The output is the solution of the QCP solver, i.e., the estimated feature $\tilde{z}$ of test data. Since the attention block in the Transformer can learn the complex relation between tokens, we train a Transformer to align the inputs and outputs of the CIP, using the solutions from the QCP solver as the training data. We use the MSE loss to fit the QCP solver's solutions. The resulting Transformer is called CIP-net. Assume we have a training set $S$, here $N_K(\tilde{x})$ is the neighborhoods set in $S$ excluded $\tilde{x}$, then we can train the CIP-net by
\begin{equation}
    \min \sum_{\tilde{x} \in S}\|\operatorname{CIP-net}(\tilde{x},N_K(\tilde{x}),Z(\tilde{x}))-\operatorname{QCP}(\tilde{x})\|_2^2
\end{equation}

When the process of solving the CIP is replaced by CIP-net, the resulting OTAD-T-NN becomes differentiable. Then the gradient with respect to inputs can be used to construct adversarial examples. However, experiments show that OTAD-T-NN remains robust, indicating that the robustness of OTAD is not only due to gradient obfuscation. As a result of that, we need to understand the robustness of a Transformer block by bounding its Lipschitz constant.

However, dot product self-attention has been proven to be not globally Lipschitz \cite{kim2021lipschitz}, which results in OTAD-T-NN not being globally Lipschitz. Nonetheless, adversarial robustness is more concerned with locally Lipschitz, meaning the Lipschitz constant within the range of adversarial examples. We provide the upper bound for the local Lipschitz constant of dot-product multihead self-attention when the input is bounded.

\begin{theorem}
Given a sequence $x_1,x_2,...,x_N \in \mathbb{R}^{D}$, the input $X = [x_1,x_2,\cdots,x_N]^T \in \mathbb{R}^{N \times D}$ is bounded by $\|X\|_F \leq M$. For $1 \leq r \leq R$, $R$ is the number of head, let $Q^{(r)}, K^{(r)}, V^{(r)} \in \mathbb{R}^{D \times D/R}$, and $W \in \mathbb{R}^{D \times D}$, assume all parameters are bounded by \[
\max_{r=1, \ldots, R} \left\{\|Q^{(r)}\|_F, \|K^{(r)}\|_F, \|V^{(r)}\|_F, \|W\|_F \right\} \leq M_{\theta}.\] 
The multi-head self-attention $F$ is defined by
$$
F(X) = [f^{1}(X),\cdots, f^{R}(X)]W,
$$
where $f^{(r)}$ is single-head dot-product self-attention defined by
\begin{equation}
f^{(r)}(X):=\operatorname{softmax}\left(\frac{X Q^{(r)} \left(X K^{(r)}\right)^{\top}}{\sqrt{D / R}}\right) X V^{(r)}.
\end{equation}
Then $F$ with bounded input is Lipschitz with the following bound on $\operatorname{Lip}_2 (F)$:
\begin{equation}
\operatorname{Lip}_2(F) \leq \sqrt{R} M_{\theta}^2 \left( \dfrac{ M_{\theta}^2}{\sqrt{D/R}} M^2 (\sqrt{N}+1) + \sqrt{N}\right) 
\end{equation}
\end{theorem}

\begin{IEEEproof}
Note that the local Lipschitz constant $\operatorname{Lip}_2(f)=\sup _{\left\|X\right\|_F \leq M}\left\|J_f(X)\right\|_2$, we can bound $\operatorname{Lip}_2(f)$ by the Jacobian $\left\|J_f\right\|_2$ with bounded input.

Let $A^{(r)}=K^{(r)}Q^{(r)\top}/\sqrt{D / R}\in \mathbb{R}^{D \times D}$, consider the map $\tilde{f}$ from $\mathbb{R}^{N \times D}$ to $\mathbb{R}^{N \times D}$, where $f^{(r)}=\tilde{f}^{(r)} V^{(r)}$,
$$
\tilde{f}:=\operatorname{softmax}\left(XA^{\top}X^{\top}\right) X 
$$
The Jacobian of $\tilde{f}$ is 
$$
J_{\tilde{f}}=\left[\begin{array}{ccc}
	J_{11} & \ldots & J_{1 N} \\
	\vdots & \ddots & \vdots \\
	J_{N 1} & \ldots & J_{N N}
\end{array}\right] \in \mathbb{R}^{N D \times N D}
$$
where \( J_{ij} = X^{\top} P^{(i)} \left[E_{ji} XA^{\top} + XA \delta_{ij}\right] + P_{ij} I \). Here, \( E_{ij} \in \mathbb{R}^{N \times N} \) is a binary matrix with zeros everywhere except at the \((i, j)\)-th entry, \(\delta_{ij}\) is the Kronecker delta, and \( P^{(i)} := \operatorname{diag}(P_{i:}) - P_{i:}^{\top} P_{i:} \). The vector \( P_{i:} \) is defined as \( \operatorname{softmax}(XA^{\top}x_i) \).

Consider the \(i\)-th row $J_i$ in $J_{\tilde{f}}$, note that $\|AB\| \leq\|A\|\|B\|,\|A+B\| \leq\|A\|+\|B\|$ and $\left\|\left[A_1, \cdots, A_N\right]\right\| \leq \sum_i\left\|A_i\right\|$,
\begin{equation*}
\begin{aligned}
& \left\| J_i \right\|_2 = 
\left\|\left[J_{i 1}, \cdots, J_{i N}\right]\right\|_2 \\
& \leq \sum_j \left\| J_{ij} \right\|_2 \\
& \leq \sum_j \left\|X^{\top} P^{(i)}E_{j i} XA^T\right\|_2+ \left\|X^{\top} P^{(i)} XA^T\right\|_2
+ \sum_j \left\|P_{ij} I\right\|_2 \\
& \leq  \left\|A\right\|_2 \left( \sum_j \left\|X^{\top} P^{(i)}E_{j i} X\right\|_2+ \left\|X^{\top} P^{(i)} X\right\|_2\right) + 1 \\
\end{aligned}
\end{equation*}

Note that $X^{\top} P^{(i)} X$ is a covariance matrix of discrete distribution $\mathbb{X}$, where $\mathbb{P}(\mathbb{X}=x_j)=P_{ij}, j = 1,...,N, \sum_i P_{ij} =1 $.
\begin{equation*}
\begin{aligned}
&\left\|X^{\top} P^{(i)} X\right\|_2 = \left\|\operatorname{Cov} \mathbb{X}\right\|_2 \leq \operatorname{Tr} \left( \operatorname{Cov} \mathbb{X}\right) \\
&= \sum_j P_{i j}\left\|x_j-\sum_k P_{i k} x_k\right\|_2^2  = \sum_j P_{i j}\left\|x_j\right\|_2^2 - \left\|\sum_k P_{i k} x_k\right\|_2^2 \\
&\leq \sum_j P_{i j}\left\|x_j\right\|_2^2 \leq \sum_j \left\|x_j\right\|_2^2 = \left\|X\right\|_F^2
\end{aligned}
\end{equation*}

By Cauchy-Schwarz inequality,
\begin{equation*}
\begin{aligned}
& \sum_j \left\|X^{\top} P^{(i)} E_{j i} X\right\|_2 =
\sum_j \left\|P_{i j}\left(x_j-\sum_k P_{i k} x_k\right) x_i^{\top}\right\|_2 \\
& = \sum_j \left( \sqrt{P_{i j}}\left\|x_j-\sum_k P_{i k} x_k\right\|_2\sqrt{P_{i j}}\left\|x_i\right\|_2\right) \\
& \leq \left( \sum_j P_{i j}\left\|x_j-\sum_k P_{i k} x_k\right\|_2^2\right) ^{1/2}\left( \sum_j P_{i j}\left\|x_i\right\|_2^2\right) ^{1/2} \\
& = \sqrt{\operatorname{Tr} \left( \operatorname{Cov} \mathbb{X}\right) } \left\|x_i\right\|_2 \leq \left\|X\right\|_F\left\|x_i\right\|_2
\end{aligned}
\end{equation*}
Thus, 
\begin{equation}
\begin{aligned}
& \left\| J_i \right\|_2 \leq \sum_j \left\| J_{ij} \right\|_2 \\
& \leq  \left\|A\right\|_2 \left( \sum_j \left\|X^{\top} P^{(i)}E_{j i} X\right\|_2+ \left\|X^{\top} P^{(i)} X\right\|_2\right) + 1 \\
& \leq \left\|A\right\|_2 \left( \left\|X\right\|_F\left\|x_i\right\|_2 +\left\|X\right\|_F^2 \right)  + 1
\end{aligned}
\end{equation}

\begin{lemma}\label{lemma1}
Let $A$ be a block matrix with block columns or rows $A_1, \ldots, A_N$. Then 
$\|A\|_2 \leq \sqrt{\sum_i\left\|A_i\right\|_2^2}$.
\end{lemma}

Given the bound of input and Cauchy-Schwarz inequality, we have $\sum_i \left\|x_i\right\|_2 \leq  \sqrt{N} \left\|X\right\|_F \leq \sqrt{N}M$, then by lemma \ref{lemma1}, 
\begin{equation}
\begin{aligned}
& \left\| J_{\tilde{f}} \right\|_2 \leq  \sqrt{\sum_i \left\| J_{i} \right\|_2^2 } \\
& \leq \sqrt{\sum_i \left(\left\|A\right\|_2 \left( \left\|X\right\|_F\left\|x_i\right\|_2 +\left\|X\right\|_F^2 \right)  + 1 \right) ^2 }\\
& 
\leq \bigg(
		\left\|A\right\|_2^2 \left\|X\right\|_F^2 \sum_i \left( \left\|x_i\right\|_2^2 + 2\left\|x_i\right\|_2\left\|X\right\|_F+ \left\|X\right\|_F^2 \right) \\
& \hspace{2em}  + 2\left\|A\right\|_2 \left\|X\right\|_F \sum_i
		\left( \left\|x_i\right\|_2 + \left\|X\right\|_F\right)  + N
\bigg)^{\frac{1}{2}} \\
& =  \bigg(
		\left\|A\right\|_2^2 \left\|X\right\|_F^4 (N+1) + 2\left\|A\right\|_2^2 \left\|X\right\|_F^3 \sum_i \left\|x_i\right\|_2 \\
& \hspace{2em}	+ 2  \left\|A\right\|_2 \left\|X\right\|_F^2 N 
		+ 2 \left\|A\right\|_2 \left\|X\right\|_F \sum_i \left\|x_i\right\|_2 + N
\bigg)^{\frac{1}{2}} \\
& \leq \sqrt{\left\|A\right\|_2^2 \left\|X\right\|_F^4 (\sqrt{N}+1)^2 + 2\left\|A\right\|_2 \left\|X\right\|_F^2(N+\sqrt{N}) + N} \\
& = \left\|A\right\|_2 \left\|X\right\|_F^2(\sqrt{N}+1) + \sqrt{N} \\
& \leq \left\|A\right\|_2 M^2 (\sqrt{N}+1) + \sqrt{N}
\end{aligned}
\end{equation}

Hence, $\operatorname{Lip}_2(\tilde{f}) \leq \dfrac{\left\|K Q^T\right\|_2}{\sqrt{D/H}} M^2 (\sqrt{N}+1) + \sqrt{N}$. For single-head dot-product self-attention $f^{(r)}=\tilde{f}^{(r)}V^{(r)}$, any parameter matrix $\|A\|_2 \leq \|A\|_F \leq M_{\theta}$, then 
\begin{equation*}
\begin{aligned}
\operatorname{Lip}_2(f^{(r)}) & \leq \left\|V^{(r)}\right\|_2 \left( \dfrac{\left\|K^{(r)} Q^{(r)T}\right\|_2}{\sqrt{D/R}} M^2 (\sqrt{N}+1) + \sqrt{N}\right) \\
& \leq M_{\theta} \left( \dfrac{ M_{\theta}^2}{\sqrt{D/R}} M^2 (\sqrt{N}+1) + \sqrt{N}\right)
\end{aligned}
\end{equation*}
Finally, for multi-head $F$, $F(X) = [f^{1}(X),\cdots, f^{R}(X)]W$, by lemma \ref{lemma1},
\begin{equation*}
\begin{aligned}
\operatorname{Lip}_2(F) & \leq\left(\sqrt{\sum_r\left\|J_{f^{(r)}}\right\|_2^2}\right)\left\|W\right\|_2 \\
& \leq
\sqrt{R} M_{\theta}^2 \left( \dfrac{ M_{\theta}^2}{\sqrt{D/R}} M^2 (\sqrt{N}+1) + \sqrt{N}\right)
\end{aligned}
\end{equation*}
\end{IEEEproof}

The theorem shows that the local Lipschitz constant of self-attention is bounded by the norm of the parameter matrix. Consequently, training with weight decay can help reduce the Lipschitz constant of self-attention, resulting in a more robust model. In our setting, the CIP-net is trained on solutions from the QCP solver. Recall the QCP problem (\ref{eq:3}), whose constraints restrict the feasible region to a small space. This makes predicting the QCP solution with CIP-net more manageable. During training, the prediction error of CIP-net quickly decreases, and the weight decay term is effectively optimized, leading to a CIP-net with a smaller Lipschitz constant and enhanced robustness in OTAD-T-NN. 

\subsection{Finding better neighbors through metric learning}
In the inference time of OTAD, the feature of test data $x^\prime$ is related to its $K$ nearest neighbors. For classification tasks, incorrect class neighbors contain a lot of obfuscated information. Therefore, finding the correct neighbors is crucial for the effectiveness of OTAD. 
Usually, we can use $l_2$ distance to find neighbors. However, $l_2$ distance is not a good metric in high dimensional space, i.e., two samples with a small $l_2$ distance may not necessarily be semantically similar. To this end, we adopt a metric learning method to discover a more suitable metric for neighbor searching.

Metric learning aims at capturing the semantic relationships between data. Classical metric learning methods learn an optimal metric from the specific data and task, such as Mahalanobis distance and its kernelizations \cite{yang2006distance}. Subsequently, the methods evolved to focus on learning a feature map $\psi(\cdot)$ so that the semantically similar samples are closer in feature space, as measured by a given specific metric like $l_2$ distance. To efficiently learn a suitable metric in a high-dimensional complex space, we introduce deep metric learning, i.e., $\psi(\cdot)$ is implemented as a DNN. For classification tasks, we optimize the deep metric learning (DML) network $\psi_\theta(\cdot)$ by a chosen loss function such as contrastive loss \cite{chopra2005learning} or triplet loss\cite{weinberger2009distance}. The optimized DML-net $\psi_\theta(\cdot)$ ensures the features from the same class are closer together while features from different classes are pushed apart, enhancing the accuracy and effectiveness of the learned metric for tasks like neighbor searching and classification.

In this paper, we choose shallow networks as 3-block ResNet or one attention block, and deep networks as ViT-B/16 in deep metric learning for neighbor searching. Our results demonstrate that deep metric learning can significantly enhance the performance of OTAD in handling complex data. However, DML-net itself remains vulnerable to adversarial attacks. For instance, an untargeted Projected Gradient Descent (PGD) attack can disrupt the learned features, making OTAD worse. In a word, we can always search for a robust DML-net $\psi_\theta(\cdot)$ \cite{robdml} for neighbor searching to improve the performance of OTAD while defending against adversarial attacks. If $\psi_\theta(\cdot)$ is not a robust model, we face a trade-off: better neighborhoods but increased vulnerability.

\section{Experimental results}
In this section, we evaluate OTAD and its extensions. For OTAD, we train it on datasets of diverse data types, e.g., image, single-cell transcriptomics, and industrial tabular data. We compare the performance of OTAD against adversarial attacks on these datasets. We change the hyperparameter in OTAD to show its effect on the Lipschitz constant.
For OTAD-T, we test it with DML-net on more complex datasets against adversarial attacks and evaluate its robustness.
For OTAD-T-NN, we show its inference time and robustness against gradient-free and gradient attacks. 
We also compare our method with the commonly used $k$ nearest neighborhood-based method (KNN). Since our model is constructed based on the theoretical understanding of residual networks \cite{gai2021mathematical}, we also test similar models based on plain networks without residual connections for comparison. Finally, we discuss the limitation of our model on synthetic data.
We have included the results under various perturbation norms as well as the computational overhead of different OTAD variants in the Appendix.

\subsection{Experimental setup}
We evaluate several OTAD variants across different backbones and solver implementations. OTAD refers to the base model built upon a ResNet backbone, while OTAD-T adopts a Transformer backbone. OTAD-T-NN denotes OTAD-T accelerated with a neural-network-based solver for the CIP problem. For neighborhood search in both OTAD-T and OTAD-T-NN, we use either a small ResNet (OTAD-T with ResNet / OTAD-T-NN with ResNet) or an attention block (OTAD-T with attention / OTAD-T-NN with attention) as the DML-net.

We consider three data types: image (MNIST \cite{MNIST}, CIFAR10 \cite{cifar10}, and ImageNet \cite{imagenet}), single-cell transcriptomics (MERFISH retina \cite{choi2023spatial}), and industrial data (red wine quality \cite{cortez2009modeling}).
For OTAD, we set the weight decay hyper-parameter to $5\times 10^{-4}$ since we empirically found that weight decay can help ResNet learn the geodesic curve. We use an SGD optimizer with 0.9 momentum on the image (MNIST), single-cell transcriptomics, and industrial data. We use the fully connected (FC) ResNet with 5, 4, and 10 ResNet blocks on the three datasets, respectively. The intermediate dimension in each block is set to be the same as the dimension of inputs. 

For OTAD-T on CIFAR10, we train a Vision Transformer (ViT) from scratch with 7 blocks and 12 heads. The input images are divided into 64 patches mapped to a 384-dimensional embedding. The model is optimized using the AdamW optimizer with a cosine annealing learning rate scheduler. On ImageNet, we choose the commonly used ViT architecture: DeiT-S \cite{touvron21a}. These models (ResNet or ViT) achieve 98.26\%, 90.29\%, and 79.2\% test accuracy on MNIST, CIFAR10, and ImageNet, respectively.

For the metric in neighborhood searching, we primarily use the $l_2$ distance. We choose shallow networks as DML-nets to obtain more similar neighbors while maintaining robustness, including 3-block ResNet, 2-layer CNN, 1-block ViT, and an attention block.
Specifically, we choose the ViT-B/16 \cite{dosovitskiy2021an} pre-trained on ImageNet as DML-net for ImageNet. The final classification layer is replaced with a linear projection followed by $L_2$ normalization, ensuring the learned features are in a unit sphere for better distance-based comparisons.
DML-nets are trained/finetuned using the triplet loss, focusing on the difference between the anchor-positive and anchor-negative distances of the learned features.

\subsection{Evaluation of attacks}
We mainly consider the $l_2$ attack in this paper as the $l_2$ distance is adopted in the construction of OTAD. Let $\epsilon$ be the $l_2$ norm of the adversarial perturbations. We also report the results under various perturbation norms in Appendix A for completeness. Since the output of OTAD is obtained by solving the QCP, we cannot take the derivative of the output with respect to the input directly as neural networks. We choose gradient-free attacks to test the models. We introduce three attacks to evaluate our OTAD model on classification tasks:

\textbf{Adaptive CW attack \cite{carlini2017adversarial} with gradient-free solver} Adaptive CW attack generates adversarial examples by solving CW loss contained defense module, and we solve this optimization problem by NGOpt in Nevergrad \cite{nevergrad}, which is a gradient-free solver. We use NGOpt with $budget = 850$ and different mutation $\sigma = 0.1, 0.15,0.2$.

\textbf{Backward Pass Differentiable Approximation (BPDA) \cite{tramer2020adaptive} composited with PGD attack \cite{madry2017towards}} BPDA is typically used when there exists an optimization loop or non-differentiable operations in defense modules. OTAD involves an optimization loop for solving the convex integration problem. Thus, the back-propagation of this part is approximated by the original network. We mainly compare OTAD with adversarial purification methods against BPDA +PGD. The number of iterations of the PGD attack is 100 on MNIST and 20 on the other datasets. For the non-purification methods, BPDA + PGD attack reduces to standard PGD attack.

\textbf{Square Attack \cite{andriushchenko2020square}} A score-based black-box attack via random search, which only accesses the inputs and outputs of models.

For the differentiable OTAD-T-NN variant, we use one of the strongest adversarial attacks, AutoAttack \cite{croce20b}, with the default hyperparameters of standard version for evaluation, including three white-box attacks (APGD-CE \cite{croce20b}, APGD-DLR \cite{croce20b}, and FAB \cite{croce20a}) and a black-box attack (Square Attack \cite{andriushchenko2020square}).

For the regression problem, we use a non-adaptive white-box attack that finds a perturbation to maximize the $l_2$ distance between the model's output and the label \cite{gupta2021adversarial}. 

For classification problems, we use standard accuracy (the performance of the defense method on clean data) and robust accuracy (the performance of the defense method on adversarial examples generated by adaptive attacks) to measure the performance. The robust accuracy is estimated on 1000 samples randomly sampled from the test set. For the regression problem, we adopt the metrics MSE, E, and SMAPE from Gupta \textit{et al.} \cite{gupta2021adversarial} to measure the regression performance where the lower, the better.

\subsection{Model comparison and settings}
We empirically compare OTAD with three categories of defense methods, i.e., adversarial training, adversarial purification, and Lipschitz networks. OTAD’s two-step process derives a Lipschitz model from a standard ResNet or ViT, so we include existing Lipschitz networks for comparison. We also evaluate OTAD against classic adversarial defense methods to demonstrate its practical performance, including adversarial training methods that may exhibit reduced robustness on unseen threats 
and adversarial purification methods that modify the forward propagation like OTAD.

For adversarial training methods, we consider the primary PGD \cite{madry2017towards} adversarial training, FGSM \cite{goodfellow2014explaining} adversarial training, TRADES \cite{zhang2019theoretically}, MART \cite{Wang2020Improving}, and AWP \cite{wu2020adversarial}. We evaluate these methods under two experimental protocols.
First, we evaluate robustness under unseen threat models. Specifically, adversarial training methods (except for the industrial data) are trained with $l_\infty$ norm perturbations and evaluated against $l_2$ attacks. Since OTAD and other methods are not trained by adversarial samples, this setting corresponds to an unseen attack scenario for all methods. The noise level of the $l_\infty$ adversarial training $\epsilon$ is set to $\epsilon=0.1$ on MNIST and $\epsilon=8/255$ on CIFAR10.
Second, we evaluate adversarial training methods under a norm-consistent setting, i.e., training and testing are conducted under the same perturbation norm (both $l_2$ norms, $\epsilon=0.5$ on CIFAR10). 
The training attacks run 40 iterations on MNIST and 20 iterations on CIFAR10. All methods are trained from the same base models in OTAD. Due to the hardness of adversarial training with ViT architecture, we also train ResNet-18 adversarially on CIFAR10 for comparison.

For adversarial purification methods, we choose median filter \cite{xu2017feature}, STL \cite{sun2019adversarial}, GAN \cite{jin2019ape}, and DiffPure \cite{DBLP:conf/icml/NieGHXVA22} to represent image pre-processing, optimization-based reconstruction, and generative model-based purification models, respectively. The classifier used in these purification methods is the same as in OTAD. When using the BPDA + PGD attack to these purification methods, we approximate the purification process with the identity map during back-propagation. 

For Lipschitz networks, we compare with $l_\infty$-distance net \cite{zhang2021towards,zhang2021boosting} and a 1-Lipschitz $l_2$ network SOC+, which incorporates skew orthogonal convolutions with certificate regularization and Householder activation \cite{singla2021improved}. The hyperparameters of these models are the same as the original papers. Specifically, the block size of SOC+ is 1 on MNIST and 3 on CIFAR10.

\subsection{OTAD is robust on diverse scenarios}
\subsubsection{Performance on MNIST}
On the MNIST dataset, we use OTAD with $K =10$, $L =2$, $l=0$, $\delta ^1 =\delta ^2 = 0.2$ to defend three adaptive attacks of different noise levels as mentioned above. 
OTAD achieves better robust accuracy in most settings (Tables \ref{tab:cw}, \ref{tab:bpda}, \ref{tab:square}), only slightly worse than SOC+ against BPDA + PGD. OTAD guarantees a small Lipschitz constant, which makes gradient-free solvers fail to find a reliable adversarial example.

\begin{table}[h]
\setlength\tabcolsep{2pt}
\begin{center}
\caption{Performance of defense methods on MNIST against adaptive CW derivative-free attack.}
\label{tab:cw}
\begin{tabular}{ccccc}
\hline \multirow{2}{*}{ Method } & \multirow{2}{*}{ Standard Acc } & \multicolumn{3}{c}{ Robust Acc } \\
\cline { 3 - 5 } & & $\sigma=0.1$ & $\sigma=0.15$ & $\sigma=0.2$ \\
\hline
PGD adversarial training    & 98.6 & 72.5 & 30.8 & 5.8\\
FGSM adversarial training   & 97.7 & 0 & 0 & 0\\\hline
Median filter & 97.6 & 1.2 & 0.1 & 0\\
STL     & 96.4 & 4.2 & 2.6 & 1.6\\
APE-GAN & 93.3 & 28.8 & 6.0 & 0.6\\\hline
$l_\infty$-dist net & 98.5&80.0&69.8&60.4\\
SOC+    & 96.0 &  75.2 & 51.6  & 23.9 \\\hline
\textbf{OTAD} & 96.3 & \textbf{81.3} & \textbf{73.4} & \textbf{63.2}\\\hline
\end{tabular}
\end{center}
\end{table}

To further investigate the ability of OTAD, we vary the hyper-parameters $L-l$ and record the standard accuracy, robust accuracy, local Lipschitz constant of OTAD, and relative error (RE). The local Lipschitz constant is computed at a noise level $\epsilon$ around $0.3$ that equals the adaptive attack. RE measures the gap between ResNet $R(\cdot)$ and OTAD feature $\tilde{f}(\cdot)$,
$$ \text{RE}(x) = \dfrac{\|R(x)-\tilde{f}(x)\|_2^2}{\|R(x)\|_2^2}$$

\begin{figure*}[ht]
	\begin{center}
	\includegraphics[width=0.92\textwidth]{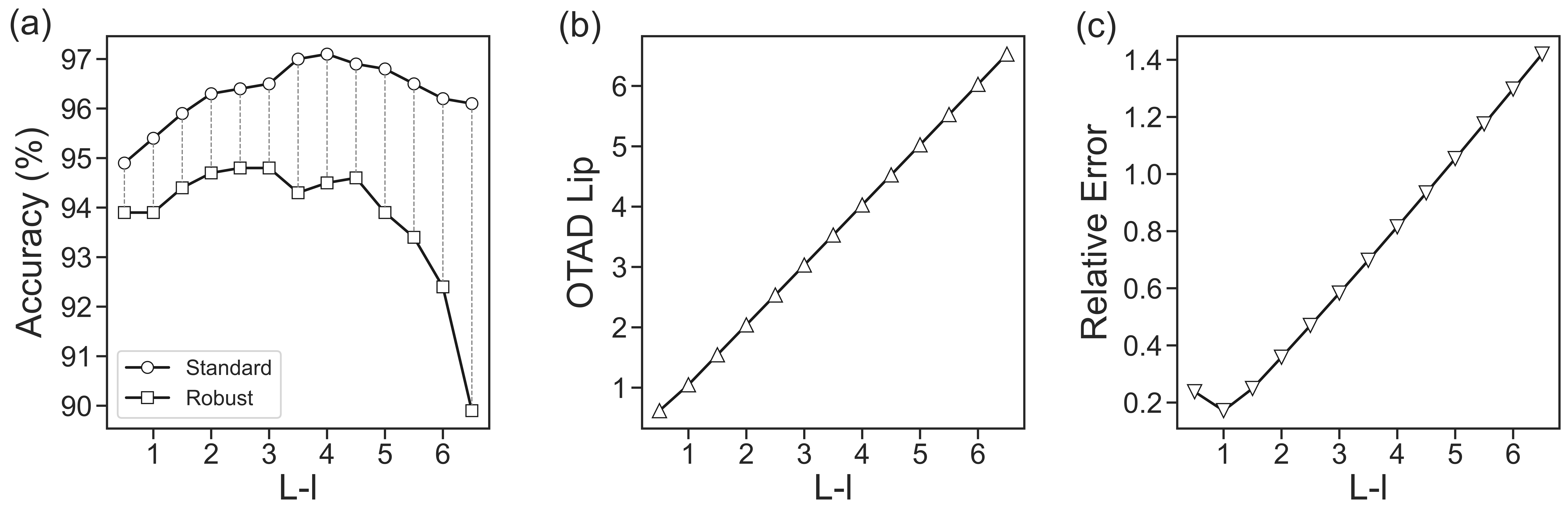}
	\caption{Performance of OTAD with different hyperparameter $L-l$ against BPDA + PGD ($\epsilon = 3$). (a) The increasing disparity between standard and robust accuracy as $L-l$ varies. 
(b) The local Lipschitz constant of OTAD (OTAD Lip) increases with $L-l$. (c) The relative error increases with increasing $L-l$. }
     \label{hyper}
     \end{center}
\end{figure*}

We can see that $L-l$ controls the local Lipschitz constant of OTAD (Fig. \ref{hyper}b). Thus, we may set smaller $L-l$ to make OTAD more robust. However, smaller $L-l$ is not always better. The local Lipschitz constant of the original ResNet is approximately $5.6$. Smaller OTAD Lip ($< 5.6$) results in better robustness (smaller disparity between standard and robust accuracy) but worse standard accuracy (Fig. \ref{hyper}a). That is because OTAD deviates from the ground truth. A larger OTAD Lip results in worse robustness, larger RE, and worse standard accuracy (Fig. \ref{hyper}c). That is because OTAD deviates from the ResNet.
The best result can be achieved with $L-l$ slightly smaller than the empirical Lipschitz constant of the original ResNet.

\begin{table}[t]
\setlength\tabcolsep{3pt}
\begin{center}
\caption{Performance of defense methods on MNIST against BPDA + PGD.}
\label{tab:bpda}
\begin{tabular}{ccccc}
\hline \multirow{2}{*}{ Method } & \multirow{2}{*}{ Standard Acc } & \multicolumn{3}{c}{ Robust Acc } \\
\cline { 3 - 5 } & & $\epsilon=2$ & $\epsilon=2.5$ & $\epsilon=3$ \\
\hline
PGD adversarial training    & 98.6 & 12.3 & 2.6 & 0.3\\
FGSM adversarial training   & 97.7 & 4.0 & 2.2 & 1.3\\\hline
Median filter & 97.6 & 0 & 0 & 0\\
STL       & 96.4 & 0 & 0 & 0\\
APE-GAN   & 93.3 & 70.3 & 48.3 & 26.8\\\hline
$l_\infty$-dist net & 98.5&78.4&74.7&71.5\\
SOC+      & 96.0 & 95.2 & \textbf{95.2} & \textbf{94.9}\\\hline
\textbf{OTAD} & 96.3 & \textbf{95.2} & 95.0 & 94.7\\\hline
\end{tabular}
\end{center} 
\end{table}

\begin{table}[t]
\begin{center}
\caption{Performance of defense methods on MNIST against  Square Attack with $2000$ queries.}
\label{tab:square}
\begin{tabular}{cccc}
\hline \multirow{2}{*}{ Method } & \multirow{2}{*}{ Standard Acc } & \multicolumn{2}{c}{ Robust Acc } \\
\cline { 3 - 4 } & & $\epsilon=2$ & $\epsilon=3$ \\
\hline
PGD adversarial training     & 98.6 & 25.6 & 1.2\\
FGSM adversarial training    & 97.7 & 0 & 0\\\hline
Median filter & 97.6 & 0 & 0\\
STL      & 96.4 & 0 & 0\\
APE-GAN  & 93.3 & 4.2 & 0.2\\\hline
$l_\infty$-dist net & 98.5&14.0&16.3\\
SOC+     & 96.0 & 56.4  & 26.1\\\hline
\textbf{OTAD} & 96.3 & \textbf{63.0} & \textbf{41.6}\\\hline
\end{tabular}
\end{center}
\end{table}

\subsubsection{Performance on the single-cell transcriptomics data}
OTAD is suitable for other data types. We use the MERFISH retina single-cell transcriptomics data by Chen et al. \cite{choi2023spatial} and preprocess it with Scanpy. The dataset is a 7-class cell-type classification problem containing 82122 single cells (65697 for training and 16425 for testing), and each cell contains expression profiles for 500 genes. Since it is hard to use generative models (adversarial purification) for single-cell transcriptomics data, we only compare OTAD with the PGD adversarial training (Table \ref{tab:TN3}). OTAD still achieves better robust accuracy than the PGD adversarial training.
\begin{table}[h]
\begin{center}
\caption{Performance of defense methods on the MERFISH retina single-cell transcriptomics data against BPDA + PGD ($\epsilon=1$).}
\label{tab:TN3}
\begin{tabular}{ccc}
\hline Method & Standard Acc & Robust Acc \\
\hline
PGD  adversarial training & 81.6 & 67.9 \\
\textbf{OTAD} & \textbf{86.6} & \textbf{73.9} \\\hline
\end{tabular}
\end{center}
\end{table}

\subsubsection{Performance on the industrial data}
In addition to classification, OTAD is also suitable for the regression task on the tabular data. We use the red wine quality dataset \cite{cortez2009modeling}, containing 1,599 samples (1279 for training and 320 for testing) with 11 features. The features are physicochemical and sensory measurements for wine. The output is a quality score ranging from 0 to 10. The testing results under different noise levels are shown in Table \ref{table:regression}. Due to DNN's robustness to random noise, the metrics computed by adding random noise can be regarded as standard. The attack makes the metrics worse, and OTAD makes the gap closer. The adversarial training is trained by adversarial examples ($\epsilon = 0.2$). OTAD achieves better results in most cases. Thus, OTAD is suitable for both classification and regression.

\begin{table*}[t!]
\begin{center}
\caption{Performance of the defense methods on the red wine quality dataset. The subscripts of these metrics are the $l_2$ norm of adversarial and random noise. Random noise indicates the Gaussian random noise added to the clean data.}
\label{table:regression}
{%
\begin{tabular}{cccc|ccc|ccc}
\hline
& $\text{MSE}_{0.1}$ & $\text{E}_{0.1}$ & $\text{SMAPE}_{0.1}$ 
& $\text{MSE}_{0.2}$ & $\text{E}_{0.2}$ & $\text{SMAPE}_{0.2}$ 
& $\text{MSE}_{0.5}$ & $\text{E}_{0.5}$ & $\text{SMAPE}_{0.5}$\\
\hline
Adversarial attack & 0.507 & 0.086 & 0.319 &
0.603 & 0.161 & 0.474 &
0.871 & 0.340 & 0.708\\
Random noise & 0.408 & 0.022 & 0.133 &
0.404 & 0.043 & 0.206 &
0.403 & 0.109 & 0.378\\
\hline
Adversarial training & 0.408 & 0.030& 0.113 &
0.440 & 0.060& 0.200 &
0.544 & 0.146& 0.381\\
\textbf{OTAD} & 0.444 &\textbf{ 0.026} &\textbf{0.079} &
0.464 &\textbf{ 0.055} &\textbf{0.188} &
\textbf{0.519} &\textbf{ 0.119} &0.383\\\hline
\end{tabular}%
}
\end{center}
\end{table*}

\subsection{OTAD-T can deal with more complex data}
For more complex datasets such as CIFAR10 and ImageNet, we test OTAD-T with DML-net for $K$ neighbor searching with parameters $K=10$, \(L=2\), \(l=0\), and \(\delta^1 = \delta^2 = 1\). As with OTAD, we first choose a gradient-free attack, BPDA + PGD. For CIFAR10, we focus on two types of DML-nets, i.e., a 3-block ResNet and an attention block.
The results show that OTAD-T with a 3-block ResNet achieves the best accuracy, and OTAD-T with an attention block demonstrates robustness but has lower standard accuracy (Table \ref{tab:otad-t-BPDA}). Additionally, OTAD-T maintains robustness against BPDA + PGD for multiple values of epsilon (Table \ref{tab:otad-t-BPDA-vary}). 

When an effective DML-net is used, OTAD-T demonstrates strong robustness against BPDA+PGD. We evaluate the adversarial training methods with both ViT and ResNet-18, which achieve comparable robust accuracy across architectures. 
However, adversarial training methods are vulnerable to unseen threats, causing a significant drop in robustness compared to OTAD-T. In the norm-consistent setting, adversarial training methods achieve higher robust accuracy than in unseen threat evaluation. Nevertheless, their performance  remains below that of OTAD-T, highlighting the superior robustness of our approach.
Adversarial purification methods using powerful generative models, such as diffusion models, also exhibit strong robustness. OTAD-T achieves competitive results when compared to DiffPure. Importantly, OTAD does not require training an additional generative model and can be directly applied to different types of data.
Lipschitz networks tend to perform modestly on more complex datasets, often resulting in lower standard accuracy. Therefore, while Lipschitz networks can be effective in certain tasks like MNIST, OTAD-T can deal with more complex data.

\begin{table}[t]
\centering
\caption{Performance of defense methods on CIFAR10 against BPDA + PGD ($\epsilon=0.5$).}
\label{tab:otad-t-BPDA}
\begin{tabular}{lcc}
\hline
Method & Standard Acc & Robust Acc \\
\hline
\multicolumn{3}{l}{\textit{Adversarial Training with $l_\infty$}} \\
PGD adversarial training & 77.7 & 63.4 \\
PGD adversarial training (ResNet-18) & 82.6 & 62.4 \\
TRADES & 84.4 & 64.3 \\
TRADES (ResNet-18) & 88.6 & 64.3 \\
MART & 71.0 & 61.3 \\
MART (ResNet-18) & 85.2 & 64.9 \\
TRADES + AWP & 84.5 & 65.4 \\
TRADES + AWP (ResNet-18) & 89.7 & 67.1 \\
\hline
\multicolumn{3}{l}{\textit{Adversarial Training with $l_2$}} \\
PGD adversarial training & 85.7 & 61.4 \\
PGD adversarial training (ResNet-18) & 89.7 & 66.3 \\
TRADES & 88.8 & 64.1 \\
TRADES (ResNet-18) & 90.2 & 70.2 \\
MART & 85.2 & 62.2 \\
MART (ResNet-18) & 89.5 & 71.6 \\
TRADES + AWP & 86.4 & 65.5 \\
TRADES + AWP (ResNet-18) & 89.4 & 74.1 \\
\hline
\multicolumn{3}{l}{\textit{Adversarial Purification}} \\
Median filter & 89.5 & 34.2 \\
STL & 87.2 & 37.8 \\
APE-GAN & 84.3 & 30.1 \\
DiffPure & 88.1 & 83.5 \\
\hline
\multicolumn{3}{l}{\textit{Lipschitz Networks}} \\
$l_\infty$-dist net & 56.1 & 20.3 \\
SOC+ & 77.8 & 45.9 \\
\hline
\multicolumn{3}{l}{\textit{Ours}} \\
\textbf{OTAD-T with attention} & 60.6 & 59.0 \\
\textbf{OTAD-T with ResNet} & \textbf{91.2} & \textbf{86.1} \\
\textbf{OTAD-T-NN with attention} & 61.9 & 60.4 \\
\textbf{OTAD-T-NN with ResNet} & \textbf{91.1} & \textbf{85.8}  \\

\hline
\end{tabular}
\end{table}

\begin{table}[h]
\setlength\tabcolsep{3pt}
\begin{center}
	\caption{Performance of OTAD-T on CIFAR10 against BPDA+PGD with different perturbation magnitudes.}
	\label{tab:otad-t-BPDA-vary}
	\begin{tabular}{ccccc}
		\hline \multirow{2}{*}{ Method } & \multirow{2}{*}{ Standard Acc } & \multicolumn{3}{c}{ Robust Acc } \\
		\cline { 3 - 5 } & & $\epsilon=0.5$ & $\epsilon=1.0$ & $\epsilon=1.5$ \\
		\hline
		OTAD-T with ResNet    & 91.2 & 86.1 & 82.9 & 79.9\\
		OTAD-T with attention   & 60.6 & 59.0 & 58.2 & 57.9\\\hline
	\end{tabular}
\end{center}
\end{table}

Introducing DML-net brings additional robustness challenges. We use an untargeted PGD attack to disrupt the learned features and test the performance of OTAD-T under this DML attack. Table \ref{tab:dmlattack} shows the robustness of OTAD-T with different DML-nets under the DML attack on CIFAR10. Despite OTAD-T with ResNet's strong defense against BPDA attacks, it can still be vulnerable to DML attacks. For a non-robust DML-net, we face a trade-off, i.e., better neighborhoods but increased vulnerability.
Moreover, a precise DML-net can even help improve the standard accuracy of OTAD-T, outperforming the original network. The cooperation of the two networks achieves better results and it is difficult to attack when DML-net is unknown.

\begin{table}[h]
\begin{center}
\caption{Performance of different DML-nets in OTAD-T on CIFAR10 against DML attack ($\epsilon=0.5$).}
\label{tab:dmlattack}
\begin{tabular}{ccc}
\hline Network architecture & Standard Acc & Robust Acc \\
\hline
3-block ResNet & 91.2 & 9.5 \\\hline
2-layer CNN    & 58.3 & 9.8 \\\hline
1-block ViT    & 49.3 & 35.8 \\\hline
An attention block & 60.6 & 49.6 \\\hline
\end{tabular}
\end{center}
\end{table}

OTAD-T requires neighborhood search in the training set to solve the CIP problem. Therefore, one of the main challenges for OTAD-T to scale to large datasets is the memory and computational cost associated with searching the training data. To address this, for large datasets like ImageNet, we may select or randomly sample a subset of data points in the inference stage to reduce memory and computational cost.  

In each class of the training data, we randomly choose subsets of varying sizes, such as 50 (base), 10 (small), 5 (tiny), and 1 (nano), for the neighborhood search process in the inference stage. This results in 50K, 10K, 5K, and 1K samples for each corresponding subset size. We choose the ViT-B/16 pre-trained on ImageNet as DML-net. Based on the results in Table \ref{tab:otad-t-BPDA-imagenet}, the size of the training subset directly impacts OTAD-T's performance, but the base version is effective enough. We can increase the subset size for better performance without being concerned about memory and computational cost. Thus, OTAD is effective on large dataset like ImageNet.

Additionally, the inference time does not significantly depend on the subset size at current scale. We decompose the OTAD process into three main components: neighborhood search, LP solver, and QCP solver. The neighborhood search step, which involves identifying K neighborhoods and feeding them into the CIP solver or CIP-net, is negligible in terms of computational cost at current scale. QCP solver is the most time-consuming step (Table \ref{tab:time_memory_comparison}), with a substantial variance in execution time across different subset sizes. The neighborhood search step may affect the inference speed of OTAD, when larger subsets are used and more neighborhoods need to identify. Nevertheless, we can randomly select a training subset to ensure that OTAD can scale to large datasets, making it suitable for large-scale applications without significant performance degradation.

\begin{table}[h]
	\centering
	\caption{Performance of OTAD-T with different sizes of subsets on ImageNet against BPDA + PGD ($\epsilon=2.0$).}
	\label{tab:otad-t-BPDA-imagenet}
	\begin{tabular}{c c c}
		\hline
		OTAD-T & Standard Acc & Robust Acc \\
		\hline
		Base  & 78.8 & 68.1  \\
		Small & 78.0 & 67.6  \\
		Tiny & 68.4 & 56.2   \\
		Nano  & 24.2 & 21.8  \\
		\hline
	\end{tabular}
\end{table}

\begin{table*}[h]
	\centering
	\caption{Comparison of average inference time (standard deviation) per sample (seconds) and memory usage (GB) for OTAD-T with different size of subsets on ImageNet across the different computation steps.}
	\label{tab:time_memory_comparison}
	\begin{tabular}{ccccc}
		\hline
		Step & Base & Small & Tiny & Nano  \\ \hline
		Neighbor search & 0.0219 s ($6.46 \times 10^{-4}$)  & 0.0172 s ($2.43 \times 10^{-4}$)  & 0.0172 s ($2.64 \times 10^{-4}$)  & 0.0208 s ($5.70 \times 10^{-4}$)  \\ \hline
		LP solver & 1.6111 s (0.6714)  & 0.5325 s (0.0895)  & 0.6235 s (0.0965)  & 1.4162 s (0.4713)  \\ \hline
		QCP solver & 9.5429 s (9.9035)  & 11.5611 s (26.7888)  & 9.1533 s (10.4927)  & 11.7192 s (25.4457)  \\ \hline
		Total & 11.1765 s (10.7814)  & 12.1116 s (26.9571)  & 9.7947 s (10.6045)  & 13.1567 s (26.2487)  \\ \hline
		Memory Usage & 62.2 GB & 12.3 GB & 8.6 GB & 6.3 GB \\ \hline
	\end{tabular}
\end{table*}

\subsection{OTAD-T-NN achieves fast inference}
We design an end-to-end 6-block Transformer encoder (CIP-net) for fast inference (OTAD-T-NN). The number of tokens in CIP-net equals $2K+1$, where $K$ ($K=10$) is the number of neighbors. Each token is mapped to a $d$-dimensional embedding ($d=2048$ on CIFAR10 and $d=4096$ on ImageNet). The training set of CIP-net is given by the solutions of the QCP solver of OTAD-T. Thus, CIP-net is a neural network designed for this specific optimization problem. OTAD-T-NN achieves a much faster inference speed. The additional computational cost is associated with training this Transformer, including making the training dataset and the training process. We compare the average inference time of OTAD-T and OTAD-T-NN per sample on CIFAR10 and ImageNet (Table \ref{tab:time}). For CIFAR10, the 3-block ResNet is chosen for DML-net, while for ImageNet, the base version of OTAD-T is used. 

\begin{table}[h]
\begin{center}
\caption{Average inference time (standard deviation) per sample on CIFAR10 and ImageNet (seconds).}
\label{tab:time}
\begin{tabular}{ccc}
\hline 	Dataset & Method & Inference time\\
\hline
\multirow{2}{*}{CIFAR10} & OTAD-T    & 1.9640 s ($1.13 \times 10^{-1}$) \\
&OTAD-T-NN & 0.0046 s ($2.66 \times 10^{-3}$) \\\hline
\multirow{2}{*}{ImageNet} & OTAD-T    & 11.1765 s ($1.08 \times 10^{1}$) \\
&OTAD-T-NN & 0.0138 s ($8.46 \times 10^{-5}$) \\\hline
\end{tabular}
\end{center}
\end{table}

Besides efficient inference, Table \ref{tab:otad-t-BPDA} shows the performance of OTAD-T-NN against BPDA + PGD remains robust. Additionally, though the CIP-net is trained on solutions from OTAD-T with ResNet, OTAD-T-NN with attention remains highly efficient.

OTAD-T-NN is now differentiable. We can use gradient-based adversarial attacks, such as AutoAttack, to test the robustness of OTAD-T-NN. Table \ref{tab:otad-t-nn-PGD} and \ref{tab:otad-t-nn-imagenet} show the performance of defense methods against AutoAttack. We do not test the STL and median filter due to the non-differentiable nature of their purification processes. Similar to the results under BPDA + PGD, OTAD-T-NN remains robust and achieves best performance against one of the strongest adversarial attacks, AutoAttack, indicating that gradient obfuscation is not the main reason for the robustness of OTAD. 

\begin{table}[t]
\centering
\caption{Performance of defense methods on CIFAR10 against AutoAttack ($\epsilon=0.5$).}
\label{tab:otad-t-nn-PGD}
\begin{tabular}{lcc}
\hline
Method & Standard Acc & Robust Acc \\
\hline
\multicolumn{3}{l}{\textit{Adversarial Training with $l_\infty$}} \\
PGD adversarial training & 77.7 & 58.1 \\
PGD adversarial training (ResNet-18) & 82.6 & 57.9 \\
TRADES & 84.4 & 61.8 \\
TRADES (ResNet-18) & 88.6 & 59.3 \\
MART & 71.0 & 54.7 \\
MART (ResNet-18) & 85.2 & 60.1 \\
TRADES + AWP & 84.5 & 62.4 \\
TRADES + AWP (ResNet-18) & 89.7 & 62.5 \\
\hline
\multicolumn{3}{l}{\textit{Adversarial Training with $l_2$}} \\
PGD adversarial training & 85.7 & 59.2 \\
PGD adversarial training (ResNet-18) & 89.7 & 63.0 \\
TRADES & 88.8 & 62.4 \\
TRADES (ResNet-18) & 90.2 & 67.0 \\
MART & 85.2 & 59.5 \\
MART (ResNet-18) & 89.5 & 68.8 \\
TRADES + AWP & 86.4 & 64.1 \\
TRADES + AWP (ResNet-18) & 89.4 & 71.7 \\
\hline
\multicolumn{3}{l}{\textit{Adversarial Purification}} \\
APE-GAN & 84.3 & 0.0 \\
DiffPure & 88.1 & 74.3 \\
\hline
\multicolumn{3}{l}{\textit{Lipschitz Networks}} \\
$l_\infty$-dist net & 56.1 & 1.0 \\
SOC+ & 77.8 & 41.8 \\
\hline
\multicolumn{3}{l}{\textit{Ours}} \\
\textbf{OTAD-T-NN with attention} & 61.9 & 31.3 \\
\textbf{OTAD-T-NN with ResNet} & \textbf{91.1} & \textbf{76.3} \\
\hline
\end{tabular}
\end{table}

\begin{table}[h]
	\centering
	\caption{Performance of OTAD-T-NN with different sizes of subsets on ImageNet against AutoAttack ($\epsilon=2.0$).}
	\label{tab:otad-t-nn-imagenet}
	\begin{tabular}{c c c}
		\hline
		OTAD-T & Standard Acc & Robust Acc  \\
		\hline
		Base  & 78.8 & 63.6   \\
		Small & 77.4 & 58.6   \\
		Tiny & 65.0 & 16.5      \\
		Nano  & 23.6 & 0.1    \\
		\hline
	\end{tabular}
\end{table}

The robustness of OTAD is due to solving the convex integration problem, resulting in a Lipschitz optimal transport map. CIP-net attempts to approximate this map and is trained with QCP solutions. If we train CIP-net with the original network's features $\{\tilde{z}_i\}_{i=1}^n$ instead of the QCP solver's solutions, OTAD-T-NN becomes vulnerable to attacks again. Now we introduce a loss function that allows training on both the QCP solver's solutions and the original network's features, controlled by the parameter $\alpha$,
\begin{equation}
\begin{aligned}
    \min \sum_{\tilde{x} \in S} &\bigg( \alpha \|\operatorname{CIP-net}(\tilde{x}) - \operatorname{QCP}(\tilde{x})\|_2^2  \\
    & + (1-\alpha) \|\operatorname{CIP-net}(\tilde{x}) - \tilde{z}\|_2^2 \bigg)
\end{aligned}
\end{equation}

We compare the robustness against AutoAttack of OTAD-T-NN trained by different values of $\alpha$ in Table \ref{tab:alpha}. As $\alpha$ decreases, the robustness of the model also decreases, indicating that OTAD-T-NN's robustness is primarily derived from the convex integration problem.

\begin{table}[t]
\begin{center}
\caption{Effect of $\alpha$ on the Robustness of OTAD-T-NN on CIFAR10 against AutoAttack ($\epsilon=0.5$).}
\label{tab:alpha}
\begin{tabular}{lcc}
\hline $\alpha$ & Standard Acc & Robust Acc \\
\hline
1 & 91.1 & 76.3 \\\hline
0.2 & 91.0 & 71.7 \\\hline
0 & 91.1 & 65.8 \\\hline
\end{tabular}
\end{center}
\end{table}

\subsection{Compared to KNN}
The KNN classifier predicts the class of a sample based on the majority class of its $K$ nearest training samples, which is robust. OTAD integrates information from neighboring points to compute a robust interpolation for the test data, functioning as a form of weighted KNN. Here, we compare OTAD and OTAD-T with KNN on MNIST and CIFAR10, respectively, while $L = 2$ in OTAD and $L = 12$ in OTAD-T. All methods search neighbors by the $l_2$ distance. For test data with \(K\) neighbors, the KNN feature of the test point is derived from the mean of the features of these \(K\) neighbors, while the OTAD feature is obtained by solving the convex integration problem. Table \ref{tab:KNN} shows the standard accuracy of these methods with different \(K\)s. Although OTAD outperforms KNN by only an average of 0.83\% on MNIST, OTAD-T outperforms KNN by an average of 6.2\% on CIFAR10 across $K=$ 5, 10, 15. That demonstrates that OTAD shows distinctly higher performance.

\subsection{Plain network-based OTAD exhibits reduced robustness} 
OTAD is designed based on DNNs with residual connections, as the forward propagation of a residual network approximates an optimal transport map. By leveraging the regularity of the optimal transport map, the DNN-induced discrete optimal transport map can be made continuous by solving the convex integration problem, resulting in a mapping with local Lipschitz continuity. What happens if we remove the residual connections? We can conjecture that the OTAD based on a plain network might exhibit reduced robustness. 
To investigate the necessity of residual connections for OTAD, we test the performance of ResNet and plain networks on relatively simple datasets, such as MNIST and Fashion MNIST. We compare the test accuracy, OTAD's accuracy on clean samples, and robust accuracy under Square Attack (Table \ref{tab:residual}).

\begin{table}[h]
\begin{center}
\caption{Standard Accuracy of KNN and OTAD on MNIST and CIFAR10}
\label{tab:KNN}
\begin{tabular}{ccccc}
\hline
Dataset & Algorithm & K=5 & K=10 & K=15 \\
\hline
\multirow{2}{*}{MNIST} & OTAD & \textbf{96.3} & \textbf{96.3} & \textbf{96.0} \\
 & KNN & 96.2 & 95.0 & 94.9 \\
\hline
\multirow{2}{*}{CIFAR10} & OTAD-T & \textbf{39.6} & \textbf{41.7} & \textbf{42.9} \\
 & KNN & 34.8 & 35.1 & 35.8 \\
\hline
\end{tabular}
\end{center}
\end{table}

We train 5-block fully-connected ResNets on both MNIST and Fashion MNIST, the plain networks have the same architecture but without residual connections. These plain networks achieve similar test accuracy to ResNets on both datasets in these experiments. On MNIST, the plain networks exhibit slightly better standard and robust accuracy than ResNets. However, on Fashion MNIST, the ResNets significantly outperform the plain networks in terms of robustness, demonstrating the advantage of residual connections in handling more complex datasets. Here we test models on simple datasets MNIST and Fashion MNIST because the ViT architecture without residual connection is hard to train and we cannot compare the robustness of it to that with the residual connection fairly.

\begin{table}[h]
\begin{center}
\caption{Performance of ResNet and plain network-based OTADs on MNIST and Fashion MNIST against Square Attack with $500$ queries ($\epsilon=3$ for MNIST and $\epsilon=2$ for Fashion MNIST).}
\label{tab:residual}
\begin{tabular}{ccccc}
\hline
Dataset & DNN & Test Acc & Standard Acc & Robust Acc  \\
\hline
\multirow{2}{*}{MNIST} & ResNet & 98.16 & 96.3 & 56.3 \\
 & Plain net & 98.49 & 97.2 & 59.3\\
\hline
\multirow{2}{*}{Fashion MNIST} & ResNet & 90.31 & 85.4 & \textbf{56.1} \\
 & Plain net & 90.41 & 86.0 & 47.6 \\
\hline
\end{tabular}
\end{center}
\end{table}

\subsection{The limitations of OTAD for complex data}
OTAD shows effectiveness on diverse types of data. However, OTAD depends on a DNN with an invariant dimension in each residual block and the neighborhoods of input. Thus, 
when this invariant-dimension DNN is ineffective and the neighborhoods of test data contain obfuscated information, OTAD will become less effective. We show this limitation with the synthetic data for classification.

We generate $50000$ training and $10000$ test data of dimension 128. The data for each class $k$ are sampling from a Gaussian random vector $\mathcal{N}(\boldsymbol{\mu}_k,\Sigma)$, where $\boldsymbol{\mu}_k\in\mathbb{R}^d$ is uniformly sampling from the unit sphere. We change the variance from $0.1$ to $0.6$ to control the difficulty of the classification task. We train a 3-block ResNet on each synthetic dataset and record test accuracy. The performance of OTAD is shown in Table \ref{tab:synthetic}. 

As the task becomes difficult, the gap (RE) becomes wider. OTAD deviates from the ResNet more, resulting in worse standard accuracy, especially in the case of std = $0.3$. That is because OTAD depends on the neighborhood of input. For a difficult task, the neighborhood contains lots of obfuscated information, hindering the performance of OTAD.

\begin{table}[h!]
\begin{center}
\caption{Performance of OTAD on the synthetic data against BPDA + PGD ($\epsilon=1$). Std means the variance of various synthetic datasets.}
\label{tab:synthetic}
\begin{tabular}{ccccc}
\hline
Std & Net test Acc & Standard Acc & Robust Acc & Relative Error \\\hline
0.1 & 100.00 & 100.00 & 100.00 & 0.0424 \\
0.2 & 100.00 & 99.44 & 75.78 & 0.1567 \\
0.3 & 99.00 & 78.69 & 35.44 & 0.4385 \\
0.4 & 89.29 & 70.52 & 20.73 & 0.4556 \\
0.5 & 70.47 & 55.42 & 14.61 & 0.4880 \\
0.6 & 51.68 & 48.28 & 12.23 & 0.5876 \\\hline
\end{tabular}
\end{center}
\end{table}

\section{Conclusion and discussion} 
DNNs are fragile to adversarial attacks. To address this issue, we developed a novel two-step model OTAD to achieve accurate training data fitting while preserving the local Lipschitz property. First, we train a DNN to obtain a discrete optimal transport map from the data to its features. Taking advantage of the regularity property inherent in the optimal transport map, we employ convex integration to interpolate the map while ensuring the local Lipschitz property. The convex integration problem can be solved through optimization solver or neural networks for fast computation. Our model suits popular architectures such as ResNets and Transformers. Experimental results demonstrate the superior performance of OTAD compared to other robust models on diverse types of datasets well-trained by a DNN with unchanged latent dimensionality. There are some directions to explore in the future:

\textbf{Defense through the cooperation of two networks}: In the section on finding better neighborhoods, we use another network to find similar neighbors as inputs to the classification network. This strategy is analogous to adversarial purification, where a generative network is adopted to purify the adversarial noise and provide clean data to the classification network. In both cases, it is hard to attack the system when the designed network is unknown. Are there other cooperation strategies for two or more networks?

\textbf{Defense using the inherent property of DNNs}: DNNs exhibit various types of implicit regularization or bias. Here, we make the most of DNNs with residual connections approximating geodesic curves in the Wasserstein space and strengthen their robustness by interpolating the discrete optimal transport map. We may find more inspiration for building robust networks from the explorations on implicit bias and make the foundation studies useful.  

\textbf{Defense based on distance between data points}: The intuition to use Euclidean distance or other distance is that the norm of adversarial noise is small compared with the norm of clean data points and the distance between data points. In this paper, we propose to find neighborhoods of test data to assist classification. But the $l_2$ distance may not find semantic similar data because high dimensional data points tend to be close to a manifold. Data points closing in $l_2$ distance may be far from each other on the manifold. How to construct a robust and accurate distance to measure the similarity between data while distinguishing data and noise is still an open question.

\section*{Acknowledgments}
This work has been supported by the Strategic Priority Research Program of the Chinese Academy of Sciences [No. XDB0680101 to S.Z.],  the CAS Project for Young Scientists in Basic Research [No. YSBR-034 to S.Z.], 
and the National Natural Science Foundation of China [Nos. 32341013, 12326614, 12126605].

\bibliographystyle{IEEEtran}
\bibliography{IEEEabrv,reference}

\vfill

\end{document}


\maketitle

\section{Results on various attacks}
Note that OTAD is not an adversarial training method: it never observes or optimizes against a particular adversarial perturbation during training. Therefore, we also evaluated its generalization to “unseen” attacks. Since in OTAD, we use $l_2$ norm to find neighbors, we treat the $l_2$ attack as a norm-consistent attack. We have added norm-consistent results for the compared models. Specifically, we report experiments where the adversarial-training baselines are trained and evaluated under the $l_2$ norm ($\epsilon=0.5$) in the main text (Tables \ref{tab:otad-t-BPDA} and \ref{tab:otad-t-nn-PGD}). We also report the results under the $l_\infty$ norm ($\epsilon=8/255$) in Tables \ref{tab:otad-t-BPDA-inf} and \ref{tab:otad-t-nn-PGD-inf}. These results demonstrate that OTAD provides competitive robustness under both norm-consistent and cross-norm settings.

To test the generalization ability of our proposed OTAD algorithm, we test it on a non-$l_p$-norm attack, Pixel \cite{Pomponi_2022}, which generates adversarial examples through sparse and structured pixel-level modifications without imposing any explicit $l_p$-norm bound on the perturbation. Results show that OTAD is superior to other models under Pixel attacks (Table \ref{tab:otad-t-pixel}). 
The Pixel attack is implemented using \textit{torchattacks} \cite{kim2020torchattacks}, with $x\_dimensions$ and $y\_dimensions$ set to $(0.03, 0.06)$, $restarts=20$, and $max\_iterations=10$.

\begin{table}[H]
\begin{center}
\caption{Performance of defense methods on CIFAR10 against BPDA + PGD $l_2$ ($\epsilon=0.5$).}
\label{tab:otad-t-BPDA}
\begin{tabular}{ccc}
\hline Method & Standard Acc & Robust Acc \\
\hline
PGD adversarial training & 85.7 & 61.4 \\
PGD adversarial training (ResNet-18) & 89.7 & 66.3  \\
TRADES & 88.8 & 64.1 \\
TRADES (ResNet-18) & 90.2 & 70.2 \\
MART & 85.2 & 62.2 \\
MART (ResNet-18) & 89.5 & 71.6 \\
TRADES + AWP & 86.4  & 65.5  \\
TRADES + AWP (ResNet-18) & 89.4 & 74.1 \\
\hline
Median filter & 89.5 & 34.2\\
STL      & 87.2 & 37.8\\
APE-GAN  & 84.3 & 30.1\\
DiffPure & 88.1 & 83.5\\\hline
$l_\infty$-dist net & 56.1 & 20.3\\
SOC+     & 77.8 & 45.9 \\\hline
\textbf{OTAD-T with attention} & 60.6 & 59.0 \\
\textbf{OTAD-T with ResNet} & \textbf{91.2} & \textbf{86.1} \\\hline
\textbf{OTAD-T-NN with attention} & 61.9 & 60.4 \\
\textbf{OTAD-T-NN with ResNet} & \textbf{91.1} & \textbf{85.8} \\\hline
\end{tabular}
\end{center}
\end{table}

\begin{table}[H]
\begin{center}
\caption{Performance of defense methods on CIFAR10 against BPDA + PGD $l_\infty$ ($\epsilon=8/255$).}
\label{tab:otad-t-BPDA-inf}
\begin{tabular}{ccc}
\hline Method & Standard Acc & Robust Acc \\
\hline
PGD adversarial training & 77.7 & 44.3 \\
PGD adversarial training (ResNet-18) & 82.6 & 45.7  \\
TRADES & 83.6 & 37.4 \\
TRADES (ResNet-18) & 85.4 & 52.5 \\
MART & 71.0 & 46.6 \\
MART (ResNet-18) & 85.2 & 51.4 \\
TRADES + AWP & 84.2  & 38.6  \\
TRADES + AWP (ResNet-18) & 86.1 & 52.6 \\
\hline
Median filter & 89.5 & 19.6\\
STL      & 87.2 & 39.2\\
APE-GAN  & 84.3 & 14.1\\
DiffPure & 87.3 & \textbf{78.2}\\\hline
$l_\infty$-dist net & 56.1 & 37.1\\
SOC+     & 77.8 & 5.8 \\\hline
\textbf{OTAD-T with ResNet} & \textbf{91.2} & \textbf{78.0} \\\hline
\textbf{OTAD-T-NN with ResNet} & \textbf{91.1} & \textbf{77.5} \\\hline
\end{tabular}
\end{center}
\end{table}
\begin{table}[H]
\begin{center}
\caption{Performance of defense methods on CIFAR10 against AutoAttack $l_2$ ($\epsilon=0.5$).}
\label{tab:otad-t-nn-PGD}
\begin{tabular}{ccc}
\hline Method & Standard Acc & Robust Acc \\
\hline
PGD adversarial training & 85.7 & 59.2 \\
PGD adversarial training (ResNet-18) & 89.7 & 63.0 \\
TRADES & 88.8 & 62.4 \\
TRADES (ResNet-18) & 90.2 & 67.0 \\
MART & 85.2 & 59.5 \\
MART (ResNet-18) & 89.5 & 68.8 \\
TRADES + AWP & 86.4 & 64.1 \\
TRADES + AWP (ResNet-18) & 89.4 & 71.7 \\
\hline
APE-GAN    & 84.3 & 0.0\\
DiffPure   & 88.1 & 74.3\\\hline
$l_\infty$-dist net & 56.1 &1.0\\
SOC+       & 77.8 & 41.8\\\hline
\textbf{OTAD-T-NN with attention} & 61.9 & 31.3 \\
\textbf{OTAD-T-NN with ResNet} & \textbf{91.1} & \textbf{76.3} \\\hline
\end{tabular}
\end{center}
\end{table}

\begin{table}[H]
\begin{center}
\caption{Performance of defense methods on CIFAR10 against AutoAttack $l_{\infty}$ ($\epsilon=8/255$).}
\label{tab:otad-t-nn-PGD-inf}
\begin{tabular}{ccc}
\hline Method & Standard Acc & Robust Acc \\
\hline
PGD adversarial training & 77.7 & 39.5 \\
PGD adversarial training (ResNet-18) & 82.6 & 42.8  \\
TRADES & 83.6 & 34.8 \\
TRADES (ResNet-18) & 85.4 & 49.0 \\
MART & 71.0 & 36.6 \\
MART (ResNet-18) & 85.2 & 47.0 \\
TRADES + AWP & 84.2  & 35.6  \\
TRADES + AWP (ResNet-18) & 86.1 & 48.4 \\
\hline
APE-GAN  & 84.3 & 0.0\\
DiffPure & 87.3 &\textbf{64.6} \\\hline
$l_\infty$-dist net & 56.1 & 26.6\\
SOC+     & 77.8 & 3.5 \\\hline
\textbf{OTAD-T-NN with ResNet} & \textbf{91.1} & \textbf{51.7} \\\hline
\end{tabular}
\end{center}
\end{table}

\begin{table}[H]
\begin{center}
\caption{Performance of defense methods on CIFAR10 against Pixel.}
\label{tab:otad-t-pixel}
\begin{tabular}{ccc}
\hline Method & Standard Acc & Robust Acc \\
\hline
PGD adversarial training & 77.7 & 39.6 \\
PGD adversarial training (ResNet-18) & 82.6 & 33.3  \\
TRADES & 83.6 & 38.3 \\
TRADES (ResNet-18) & 85.4 & 39.9 \\
MART & 71.0 & 43.1 \\
MART (ResNet-18) & 85.2 & 37.2 \\
TRADES + AWP & 84.2  & 38.5  \\
TRADES + AWP (ResNet-18) & 86.1 & 40.4 \\
\hline
Median filter & 89.5 & 22.0\\
APE-GAN  & 84.3 & 21.0\\
DiffPure & 87.3 & \textbf{62.1} \\\hline
$l_\infty$-dist net & 56.1 & 19.3 \\
SOC+     & 77.8 & 32.0 \\\hline
\textbf{OTAD-T with ResNet} & \textbf{91.2} & \textbf{66.7} \\\hline
\textbf{OTAD-T-NN with ResNet} & \textbf{91.1} & \textbf{58.5} \\\hline
\end{tabular}
\end{center}
\end{table}

\section{Computational costs of OTAD}
Our OTAD algorithm shares the same training procedure as the standard backbone classifier. We compare the computational costs between the standard backbone model and OTAD-based models, specifically OTAD-T-NN with ResNet, and OTAD-T with ResNet. In the training process, OTAD-T-NN with ResNet requires training two additional components: (i) a CIP-net to solve the convex integration problem (CIP), and (ii) a DML-net to identify better neighbors. In contrast, OTAD-T with ResNet only requires training the DML-net, while standard backbone training involves only the backbone classifier.
Since these components are trained independently and can be flexibly combined, we report the component-wise training costs, including CIP-net training, DML-net training, and standard backbone training. The overall training cost of each model is obtained by summing the corresponding components. The results are summarized in Table \ref{tab:training_cost}.

In the inference process, OTAD-T with ResNet is evaluated using a batch size of 1 due to its per-sample QCP solving procedure. In contrast, OTAD-T-NN with ResNet and the standard model are evaluated using a batch size of 100.
For each method, we report the average inference time per batch, together with the standard deviation, measured over multiple runs. The results are summarized in Table \ref{tab:inference_cost}.

The results indicate that the costs of OTAD-T and OTAD-NN are approximately 2 to 5 times higher than those of a standard classification network during both training and testing. Since it provides more robust outputs, this increase is acceptable for most applications.

\begin{table}[H]
\setlength{\tabcolsep}{3pt}
\centering
\caption{Component-wise training cost of OTAD variants and standard training on CIFAR10.}
\label{tab:training_cost}
\begin{tabular}{lccc}
\hline
\textbf{Training} 
& \textbf{CIP-net training} 
& \textbf{DML-net training} 
& \textbf{Standard training} \\
\hline
CPU Memory      & 54.8 GB  & 17.5 GB & 15.6 GB \\
GPU Memory    & 9.4 GB   & 1.2 GB & 3.5 GB \\
Training Time & 4.44 h  & 0.42 h & 2.96 h \\
Epochs               & 200   & 200  & 200 \\
\# Parameters     & 191.25M & 2.8M & 6.3M\\
Batch Size           & 512   & 256  & 128 \\
\hline
\end{tabular}
\end{table}

\begin{table}[H] 
\setlength{\tabcolsep}{3pt} 
\centering 
\caption{Inference cost comparison of OTAD variants and standard training. Inference time is reported as mean and standard deviation.} 
\label{tab:inference_cost} 
\begin{tabular}{lccc} 
\hline 
\textbf{Inference} & \textbf{OTAD-T-NN} & \textbf{OTAD-T } & \textbf{Standard network} \\ \hline 
CPU Memory & 23.8 GB & 50.9 GB & 11.6 GB \\ 
GPU Memory & 2.03 GB & 0.65 GB & 0.54 GB \\ 
Inference Time Mean & 0.0128 s  & 1.9640 s  & 0.0079 s  \\ 
Inference Time Std  & $8.0 \times 10^{-4}$  & $1.13 \times 10^{-1}$  & $2.7 \times 10^{-4}$ \\
Batch Size & 100 & 1 & 100 \\ 
\hline 
\end{tabular} 
\end{table}

\section{Remark on the zero robust accuracy of FGSM adversarial training in Table I-III}
The clean accuracy of the FGSM adversarially trained model on MNIST was relatively low (around 72$\%$), suggesting suboptimal convergence. To address this issue, we introduced a warm-up training strategy, where the perturbation strength was gradually increased from zero to $\epsilon = 0.1$. This significantly improved the clean accuracy of FGSM adversarial training to 97.7$\%$, confirming that the previous low clean accuracy was mainly due to training instability rather than overfitting. 

Nevertheless, even after stabilizing training, the robust accuracy remains extremely limited. Specifically, we only observed a slight improvement under BPDA+PGD, while the robust accuracies under the CW attacks and Square attacks remain zero, as shown in Tables I-III. This behavior suggests that the lack of robustness is not due to overfitting, but instead reflects the intrinsic limitations of the FGSM adversarial training.

The adversarial training in Table I employs FGSM with $L_\infty$perturbations, while the evaluation includes CW attacks under the $L_2$ norm, which follow a fundamentally different threat model and are considerably stronger. Therefore, it is expected that models trained with $L_\infty$-bounded FGSM adversaries exhibit low or even zero robust accuracy under $L_2$-based CW attacks. This observation is consistent with prior works on adversarial training and cross-norm generalization. For example, \cite{10.1007/978-3-030-88013-2_3} explicitly identifies that single-step training often learns perturbations that do not generalize to stronger attacks, causing robust accuracy against stronger multi-step methods to drop to zero.

\bibliographystyle{IEEEtran}
\bibliography{IEEEabrv,reference}